\makeatletter
\newcommand*{\addFileDependency}[1]{
  \typeout{(#1)}
  \@addtofilelist{#1}
  \IfFileExists{#1}{}{\typeout{No file #1.}}
}
\makeatother

\documentclass[10pt,twocolumn,letterpaper]{article}

\usepackage[pagenumbers]{cvpr} 

\usepackage{graphicx}
\usepackage{amsmath}
\usepackage{amssymb}
\usepackage{booktabs}
\usepackage{xr}
\usepackage{caption}
\usepackage{subcaption}

\usepackage{multirow}
\usepackage{xr}
\usepackage{caption}
\usepackage{subcaption}
\usepackage{booktabs}
\usepackage[export]{adjustbox}
\usepackage{colortbl}
\usepackage{float}

\usepackage{hyperref}

\usepackage[capitalize]{cleveref}
\crefname{section}{Sec.}{Secs.}
\Crefname{section}{Section}{Sections}
\Crefname{table}{Table}{Tables}
\crefname{table}{Tab.}{Tabs.}

\def\cvprPaperID{9880} 
\def\confName{CVPR}
\def\confYear{2022}

\usepackage{overpic}
\usepackage{enumitem} 
\usepackage{overpic} 
\usepackage{color}

\definecolor{turquoise}{cmyk}{0.65,0,0.1,0.3}
\definecolor{purple}{rgb}{0.65,0,0.65}
\definecolor{dark_green}{rgb}{0, 0.5, 0}
\definecolor{orange}{rgb}{0.8, 0.6, 0.2}
\definecolor{red}{rgb}{0.8, 0.2, 0.2}
\definecolor{darkred}{rgb}{0.6, 0.1, 0.05}
\definecolor{blueish}{rgb}{0.0, 0.3, .6}
\definecolor{light_gray}{rgb}{0.7, 0.7, .7}
\definecolor{pink}{rgb}{1, 0, 1}
\definecolor{greyblue}{rgb}{0.25, 0.25, 1}

\usepackage{blindtext}

\renewcommand{\paragraph}[1]{\vspace{1em}\noindent\textbf{#1}.}

\usepackage{multirow}
\usepackage{xr}
\usepackage{caption}
\usepackage{subcaption}
\usepackage{booktabs}
\usepackage{adjustbox}
\usepackage{colortbl}
\usepackage{float}
\begin{document}
\title{Medical Image Registration via Neural Fields}
\author{Shanlin Sun\\
University of California, Irvine\\
{\tt\small shanlins@uci.edu}
\and
Kun Han\\
University of California, Irvine\\
{\tt\small khan7@uci.edu}
\and
Chenyu You\\
Yale University\\
{\tt\small chenyu.you@yale.edu}
\and
Hao Tang, Deying Kong, Junayed Naushad, Xiangyi Yan, Haoyu Ma, Pooya Khosravi\\
University of California, Irvine\\
{\tt\small \{htang6, deyingk, jnaushad, xiangyy4, haoyum3, pooyak\}@uci.edu}
\and
James S. Duncan\\
Yale University\\
{\tt\small james.duncan@yale.edu}
\and
Xiaohui Xie\\
University of California, Irvine\\
{\tt\small xhx@uci.edu}
}


 
\maketitle

\begin{abstract}
Image registration is an essential step in many medical image analysis tasks. 
Traditional methods for image registration are primarily optimization-driven, finding the optimal deformations that maximize the similarity between two images. 
%
%
Recent learning-based methods, trained to directly predict transformations between two images, run much faster, but suffer from performance deficiencies due to model generalization and the inefficiency in handling individual image specific deformations.  
%
%
%
Here we present a new neural net based image registration framework, called NIR (Neural Image Registration), which is based on optimization but utilizes deep neural nets to model deformations between image pairs. 
NIR represents the transformation between two images with a continuous function implemented via neural fields, receiving a 3D coordinate as input and outputting the corresponding deformation vector.
NIR provides two ways of generating deformation field: directly output a displacement vector field for general deformable registration, or output a velocity vector field and integrate the velocity field to derive the deformation field for diffeomorphic image registration.
The optimal registration is discovered by updating the parameters of the neural field via stochastic gradient descent. We describe several design choices that facilitate model optimization, including coordinate encoding, sinusoidal activation, coordinate sampling, and intensity sampling. 
 %
%
Experiments on two 3D MR brain scan datasets demonstrate that NIR yields state-of-the-art performance in terms of both registration accuracy and regularity, while running significantly faster than traditional optimization-based methods.
%
%

\end{abstract}

\section{Introduction}


3D image registration has a pivotal role in many medical applications \cite{incoronato2017radiogenomic, risholm2011multimodal}, such as merging images from different modalities, motion correction, tracking disease progression, and atlas-based image segmentation.
Image registration can be categorized into two groups: rigid and non-rigid. Non-rigid registration (also known as deformable registration), considering non-affine coordinate transformations between two images, is more widely used. 
%
Diffeomorphic image registration, imposing additional transformation constraints, such as smoothness, invertibility and topology preservation, is often preferred in certain applications. In this paper, we present a new image registration framework that supports both general deformable and specific diffeomorphic image registrations. 


Traditional image registration methods \cite{bajcsy1989multiresolution, shen2002hammer, rueckert1999nonrigid, beg2005computing, avants2008symmetric} treat image registration as an optimization problem: finding the optimal coordinate transformations that maximize the similarity between the transformed source image and the target image.
These methods usually require hard modeling assumptions on the types of permissible deformations to ensure registration regularity. For instance, NiftyReg \cite{rueckert1999nonrigid} models deformation fields using B-splines with a set of control points. Flow-based methods model the transformations via a series of time-dependent velocity fields \cite{beg2005computing,zhang2017frequency} or stationary velocity fields \cite{avants2008symmetric}, and impose strong assumptions on the space of permissible velocity vector fields. 
The strong modeling assumptions produce well-behaved transformations, but sometimes also lead to detrimental registration outcomes.  To improve optimization-based registration requires a more flexible framework for modeling the space of permissible transformations. In addition, optimization-based methods are often time-consuming. 


Recent advances in deep learning have inspired the development of learning based image registration methods \cite{balakrishnan2019voxelmorph,dalca2018unsupervised,mok2020fast,mok2020large,mok2022affine}. 
The learning-based registration methods are trained to directly output transformations between two images. Although training may take time, predictions are usually generated through a feed-forward model and therefore are very fast.  
 However, in terms of registration accuracy, learning-based methods still lag behind the optimization-based ones under unsupervised settings, even with  very complex and large-scale network structures utilized in recent works \cite{chen2021vit, mok2022affine, shi2022xmorpher}. 
Part of the reason is due to the discrepancy between the performances of the models on training data vs.~test data.  Benefiting from high representational capacity of deep neural networks,  learning-based methods can generate high quality transformations between training image pairs, but often generalize poorly on previously unseen image pairs.
Inadequacies in size and diversity of medical datasets accentuate the generalizability issue.
%
%
To alleviate the generalizability issue, several recent works \cite{hering2021learn2reg,siebert2021fast,hager2020variable,zhu2021test} have advocated a two-step approach - using learning models to derive an initial registration, followed by traditional optimization methods for fine-tuning. 


Naturally, we ask: Can the optimization-based registration also leverage the expressive power of deep neural nets? To this end, we propose \textbf{NIR} (stands for Neural Image Registration), an optimization-based framework that solves medical image registration via neural fields. Neural fields are a class of neural networks, also called coordinate-based neural multilayer perceptrons~(MLPs) or implicit neural representation~(INR), that map a point in space and time to a continuous quantity. Previously we demonstrated the effectiveness of neural fields in modeling diffeomorphic transformations for anatomic shape analysis \cite{sun2022topology}. This motivates us to apply neural fields to model deformable and diffeomorphic registrations between images. 
NIR provides two ways of modeling image deformations, either directly modeling the displacement vector field or modeling the velocity vector field. In both cases, the neural field within NIR takes as input a 3D coordinate of the source image and outputs a 3D vector (either displacement or velocity) at the location. In the second case, the velocity vector field is further integrated through a Neural ODE Solver \cite{chen2018neuralode} to produce the final deformation field, thereby ensuring that the resulting deformation is diffeomorphic.  

Modeling deformation fields as coordinate-based MLPs, supplemented with additional features such as Fourier position encoding \cite{MatthewTancik2020FourierFL} and periodic activation functions \cite{sitzmann2020implicit} in NIR, offers several advantages. First, the neural deformation model is simple and flexible, and yet still has great expressive power. It can use a relatively small number of coefficients to encode signals with an exponentially large frequency support \cite{yuce2022structured}. Deformations with high frequencies can be captured by scaling up the number of hidden layers and neurons. Second, different from other neural nets defined on discrete grid coordinates like convolutional neural networks (CNNs), coordinate-based MLPs are defined on the continuous coordinate space. Neural fields can be optimized to model fine deformations with sampled data points and does not require dense input. Consequently, optimizing neural fields is memory-efficient. Third, the optimization of neural fields can take full advantages of the modern high-performance automatic differentiation toolboxes such as PyTorch \cite{paszke2019pytorch}, Tensorflow \cite{abadi2016tensorflow} and JAX \cite{jax2018github}.

We use stochastic gradient descent (SGD) to find the optimal parameters of the neural field in NIR and design efficient coordinates sampling strategies to run SGD. 
%
Taking registration accuracy, registration regularity, as well as convergence rate into account, two coordinate samplers --- downsize sampler and mini-patch sampler, are examined. Downsize coordinate sampler offers faster convergence and higher registration accuracy, whereas mini-patch coordinate sampler is more beneficial to the regularity of deformation fields. To bring together the strengths of these two coordinate samplers, we further propose a hybrid sampler for NIR, comprising two concatenated neural fields optimized with a downsize and a mini-patch coordinate sampler separately. Our experiments show that NIR with a hybrid sampler can perform well in both registration accuracy and regularity without significantly compromising optimization efficiency. 

The main contributions of our work are summarized as follows: 
\begin{itemize}

    \item We introduce NIR, a novel optimization-based deformable image registration framework that models the displacement field or velocity field via lightweight coordinate-based MLPs with Fourier position encoding and sinusoidal activation functions. 
    \item We further propose a hybrid sampling scheme, composed of two stacked neural fields, separately optimized with two different coordinate samplers, to efficiently solve optimization in NIR via SGD.  
    
    \item NIR is evaluated on two brain MRI datasets and shows state-of-the-art registration results in multiple metrics, including intensity similarity between fixed and transformed images, regularity of the transformation, and GPU consumption. It runs significantly faster than traditional optimization based methods, while requiring less memory than learning based methods
    (less than 3500MB GPU memory).
    
    
    
\end{itemize}


\section{Related Works}

\subsection{Optimization-based Registration Methods}

Extensive works have been conducted in 3D deformable image registration through the decades \cite{ashburner2007fast,avants2008symmetric,bajcsy1989multiresolution,beg2005computing,dalca2016patch}. Several studies solve the task of image registration as an optimization problem in the space of displacement vector fields. They optimize the deformable model iteratively with the constraint from a smoothness regularizer which is typically a Gaussian smooth filtering. These include elastic-type models \cite{bajcsy1989multiresolution}, free-form deformation with B-splines \cite{rueckert1999nonrigid}, statistic parametric mapping \cite{ashburner2000voxel}, local affine models \cite{hellier2001hierarchical} and Demons \cite{thirion1998image}. Diffeomorphic image registration with the attributes of topology preserving and transformation invertibility also achieve remarkable progress in various anatomical studies. Some of the popular methods include Large Diffeomorphic Distance Metric Mapping (LDDMM) \cite{beg2005computing}, DARTEL \cite{ashburner2007fast} and standard symmetric normalization (SyN) \cite{avants2008symmetric}. In this field, the deformation is modeled by integrating its velocity over time according to the Lagrange transport equation \cite{christensen1996deformable,dupuis1998variational} to achieve a global one-to-one smooth and continuous mapping.

\subsection{Learning-based Registration Methods}

Many learning-based methods \cite{balakrishnan2018unsupervised,dalca2018unsupervised,sheikhjafari2018unsupervised,mok2021conditional,mok2020large,mok2020fast,zhang2021learning,chen2021vit,mok2022affine,hu2021end,shi2022xmorpher} are proposed to provide promising registration results with fast inference speed and high registration accuracy. By learning the representation of images through large amount of training data, the neural networks are able to capture the difference between input pair of images and predict the transformation. VoxelMorph \cite{balakrishnan2018unsupervised} utilizes the UNet-like structure to directly regress the deformation fields by minimizing the dissimilarity between input and target images. SYM\_Net \cite{mok2020fast} provides a symmetric registration method which estimates the forward and backward deformation simultaneously within the space of the diffeomorphic maps. LapIRN \cite{mok2020large} avoids the local minima of registration in a coarse-to-fine fashion. A recursive cascaded network \cite{zhao2019recursive} was proposed to iteratively apply the registration network to the warped moving image and fixed image. DTN \cite{zhang2021learning} deploys a transformer over the CNN backbone to capture the semantic contextual relevance and enhance the extracted feature from backbone. MS-ODENet \cite{xu2021multi} chooses to learn a registration optimizer via a multi-scale neural ODE model and proposes the cross-model similarity metric to alleviate the appearance difference in different contrast levels. 

\subsection{Neural Fields for Visual Computing}

Recently, neural fields have advanced as a popular technique in solving visual computing problems. It uses coordinate-based neural networks to parameterize the physical properties of scenes and objects across space and time \cite{xie2021neural}. Initially, neural fields were designed to solve the shape representation problem \cite{park2019deepsdf,niemeyer2019occupancy}. From then on, neural fields have been applied in more computer vision tasks. Neural Radial Field introduced in \cite{mildenhall2020nerf} is designed to achieve view-dependent scene representation. Periodic sinusoidal activation function \cite{sitzmann2020implicit} are proposed to replace the relu-based activation functions for the better representation of complex natural signals. Coin \cite{dupont2021coin} compress an image by storing the weights of a neural field overfitted to it.

\subsubsection{Deformation Representation} Neural Fields can be used to represent continuous transformation with more flexibility. As target geometry and appearance are often modeled with neural fields, it is natural to use neural field to represent the transformation. \cite{niemeyer2019occupancy} performs 4D reconstruction via learned temporal and spatially continuous vector field. Neural Mesh Flow \cite{gupta2020neural} focuses on generating manifold mesh from images or point clouds via conditional continuous diffeomorphic flow. PointFlow \cite{yang2019pointflow} incorporates continuous normalizing flows with a principle probabilistic framework to reconstruct 3d point clouds. DiT \cite{zheng2021deep} builds up the dense correspondence across shapes in one category by decomposing DeepSDF \cite{park2019deepsdf} into a deformation network and a single shape representation network.

\subsubsection{Medical Imaging Application} Neural fields have been applied in some medical image analysis tasks, such as 3D image reconstruction or representation. \cite{sun2021coil} tries to augment the quantities measured in the sensor domain and reconstructs images with less measurement noise. \cite{shen2021nerp} predicts the density value at a 3D spatial coordinate, and is supervised by mapping its value back to the sensor domain. \cite{wu2021irem} views the 2D slice as the samples from 3D continuous function and reconstructs 3D images from the observed tissue anatomy. NDF \cite{sun2022topology} follows the paradigm of DiT and proposes to model the topology preserving transformation between each organ shape instance and the learned shape template via neural diffeomorphic flow.

Two recent independent works, IDIR \cite{wolterink2021implicit} and NODEO \cite{wu2022nodeo} also proposed optimization-based pair-wise image registration methods utilizing coordinate-based neural networks. IDIR is a direct extension of SIREN \cite{sitzmann2020implicit} predicting the displacement vectors of randomly sampled query coordinates during optimization. Similar to our work, NODEO also leverages Neural ODE \cite{chen2018neuralode} to integrate the velocity fields to obtain the deformation fields. However, NODEO uses a completely different network architecture. Their neural velocity field is based on a Unet-like 3D CNN model with fully connected bottleneck layers, whereas ours is a simple MLP with coordinate encoding and sinusoidal activation functions. During optimization, NODEO receives the whole grid coordinates as input and predicts the entire deformations in every iteration. Thus, NODEO requires a large memory footprint. To reduce memory consumption, NODEO must reduce the spatial size or the channel number of the feature maps, making it difficult to represent the fine deformations.


\section{Method}
\label{sec:method}

\subsection{Background}
\label{subsection:background}
\subsubsection{Pairwise Image Registration}
\label{subsubsection:registration_bg}
Let $\boldsymbol{T} \in \mathbb{R}^{D \times H \times W}$ and $\boldsymbol{M} \in \mathbb{R}^{D \times H \times W}$ denote the target and moving volumetric images, respectively. Let $\phi:\Omega\subset\mathbb{R}^3\to\Omega$ be the deformation field between $\boldsymbol{T}$ and $\boldsymbol{M}$. 
The unsupervised image registration is commonly formulated as an optimization problem:
\begin{equation}
   	\hat\phi = \arg \min_{\phi}{\mathcal{L}(\boldsymbol{T},~\boldsymbol{M},~\phi)}, \label{eq:opt_problem_1} 
\end{equation}
where the cost function
\begin{equation}
{\mathcal{L}(\boldsymbol{T},~\boldsymbol{M},~\phi)}
    = {\mathcal{L}_{sim}(\boldsymbol{T},~\boldsymbol{M}\circ \phi)+\lambda_{reg} \cdot \mathcal{L}_{reg}(\phi)}, \label{eq:opt_problem_2}
\end{equation}
includes two terms: a) $\mathcal{L}_{sim}$, measuring image similarity between the target and warped moving volumes, and b) $\mathcal{L}_{reg}$, a regularization term on the deformation field.  $\boldsymbol{M}\circ \phi$ denotes $\boldsymbol{M}$ warped by the deformation field $\phi$. $\lambda_{reg}$ is a hyperparameter controlling the relative weight of the regularization term .

Registration field $\phi$ is represented either directly via a displacement field $\mathbf{u}$ with $\phi = \mathbf{Id} + \mathbf{u}$, where $\mathbf{Id}$ is the identity map~\cite{bajcsy1989multiresolution, balakrishnan2019voxelmorph}, or indirectly via a velocity vector field $\mathbf{v}$, the integration of which leads to $\phi$. The second approach is preferred if we require the registration field to be diffeomorphic, i.e., invertible and topology preserving~\cite{beg2005computing, mok2020fast}.

\subsubsection{Neural Fields}
\label{subsubsection:neural_field}
Both displacement fields and vector fields are modeled by a coordinate-based neural net, referred to as neural field  $\mathcal{F}_{\theta}:\mathbb{R}^3\to\mathbb{R}^3$, which provides a continuous mapping from 3D coordinate $\boldsymbol{p}$ to the displacement or velocity vector at that position. $\boldsymbol{\theta}$ denotes the parameters of the neural net. 
%
%
Neural fields provide a flexible framework for modeling registration field, powerful enough to model highly complex deformations, while maintaining analytic differentiability and allowing us to leverage powerful optimization tools in existing deep learning toolboxes~\cite{frankle2018lottery}.

The neural fields used in this work all consist of a coordinate encoding layer $\gamma$, followed by a multilayer perceptron (MLP) whose weights, bias and activation function at the $\ell$-th layer are denoted as $\boldsymbol{W}^{(\ell)}$, $\boldsymbol{b}^{(\ell)}$ and $\rho^{(\ell)}$, respectively. The activities of neurons at each layer are computed sequentially as follows,
\begin{align}
\label{eq:neural_fields}
\boldsymbol{z}^{(0)} &=\gamma(\boldsymbol{p}) \nonumber \\
\boldsymbol{z}^{(\ell)} &=\rho^{(\ell)}\left(\boldsymbol{W}^{(\ell)} \boldsymbol{z}^{(\ell-1)}+\boldsymbol{b}^{(\ell)}\right),\ \ell=1, \ldots, L-1   \nonumber\\ 
\mathcal{F}_{\boldsymbol{\theta}}(\boldsymbol{p}) &=\boldsymbol{W}^{(L)} \boldsymbol{z}^{(L-1)}+\boldsymbol{b}^{(L)},
\end{align}
where $\boldsymbol{p}$ is the input coordinate and $\mathcal{F}_{\boldsymbol{\theta}}(\boldsymbol{p}) $ denotes the output displacement vector or velocity vector at $\boldsymbol{p}$.
\subsection{Overview of NIR}
NIR uses neural fields to represent the transformation between two medical images. It solves the image registration problem Eq.~(\ref{eq:opt_problem_1}) by optimizing $\boldsymbol{\theta}$. The optimization is solved via stochastic gradient descent by finding a stochastic approximation of the objective function Eq.~(\ref{eq:opt_problem_2}) through sampling, as opposed to batch gradient descent, which requires a complete calculation of Eq.~(\ref{eq:opt_problem_2}) and therefore is both memory-demanding and less efficient. 

\begin{figure}
     \centering
     \begin{subfigure}[b]{0.45\textwidth}
         \centering
         \includegraphics[height=0.35\textwidth]{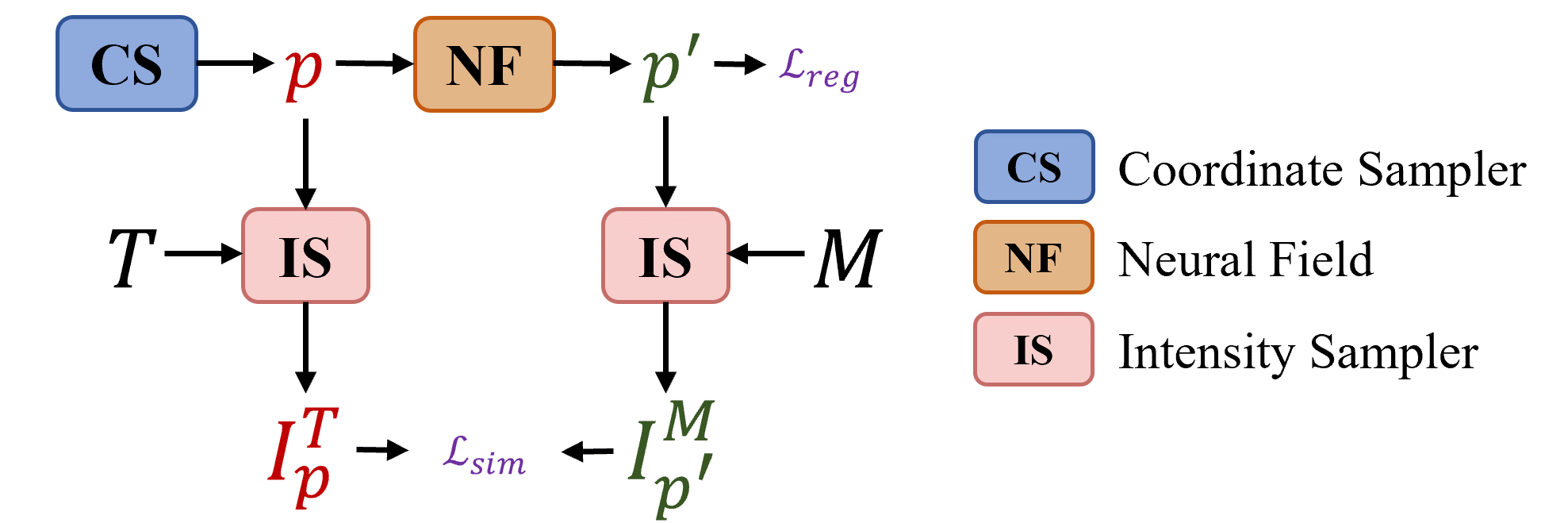}
         \caption{Optimization}
         \label{fig:single-layer-framework-opt}
     \end{subfigure}
     \hfill
     \begin{subfigure}[b]{0.45\textwidth}
         \centering
         \includegraphics[height=0.35\textwidth]{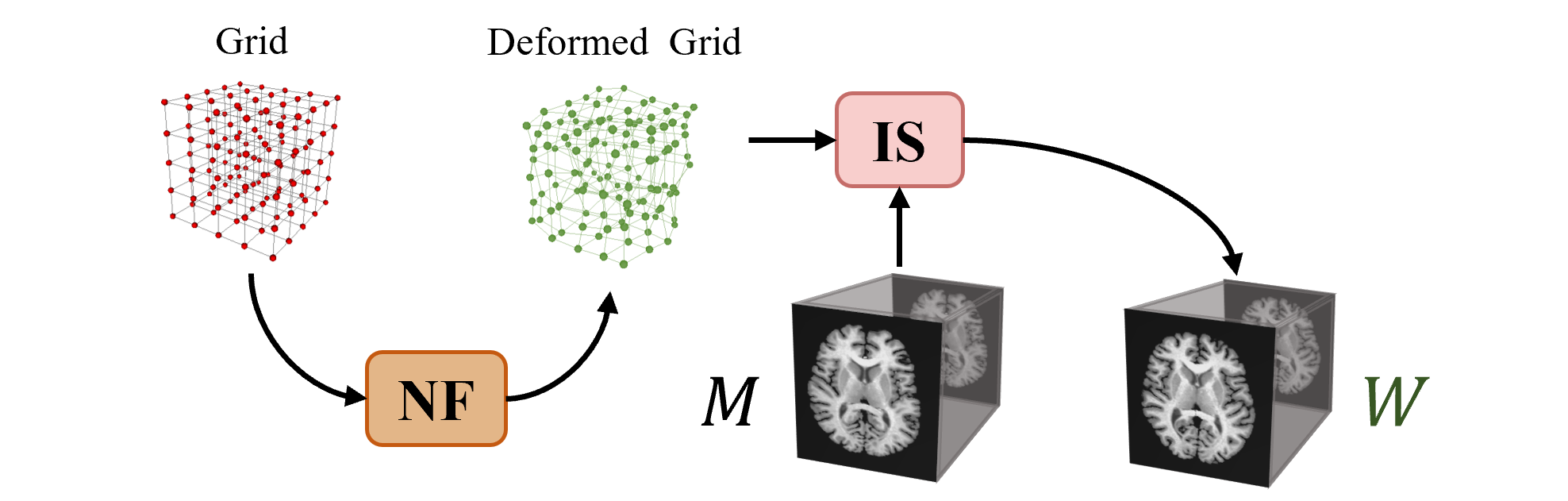}
         \caption{Inference}
         \label{fig:single-layer-framework-inf}
     \end{subfigure}
\caption{\textbf{Overview of NIR}, which is a optimization-based pairwise medical image registration framework via neural fields. In each iteration of optimization, every position $p$ is sampled from the coordinate of target volume $T$ and the deformed position $p^{\prime}$ is predicted by \textit{NF}. The intensity similarity loss $\mathcal{L}_{sim}$ between sampled image intensities is govern by the local normalized cross-correlation and the regularization term $\mathcal{L}_{jedt}$ penalizes the regions where the local deformation orientations are inconsistent, as formulated in Eq.~\ref{eq:loss}. During inference, the neural field takes as input the whole grid and outputs the deformed positions of the whole grid. Then, by sampling the intensity of the deformed grid on the moving volumes, we can get the warped volumes. Plot~(b) only presents the transformation of moving volumes via NIR, but the structures associated with the moving volumes can also be transformed in the same way.}
\label{fig:model_overview}
\end{figure}

NIR consists of three main components -- Coordinate Sampler (\textbf{\textit{CS}}), Neural Field (\textbf{\textit{NF}}), and Intensity Sampler (\textbf{\textit{IS}}) (Fig.~\ref{fig:model_overview}). 
\textit{CS} samples coordinates from the 3D grid points of $T$, randomly at each step of the optimization. 
The sampled points are sent to \textit{NF}, which maps position $\boldsymbol{p} \in \mathbb{R}^3$ in the coordinate space of $T$ to position $\boldsymbol{p^{\prime}} \in \mathbb{R}^3$ in the coordinate space of $M$.
\textit{IS} returns image intensities at query locations of source and target images. Let $I_{p}^{T}$ denote the intensity of $\boldsymbol{p}$ on $T$ and $I_{p^{\prime}}^{M}$ denote the intensity of $\boldsymbol{p}^{\prime}$ on $M$. 
The sampled image intensities are then used to calculate the similarity loss $\mathcal{L}_{sim}$ (e.g., local normalized cross-correlation loss) between $I_{p}^{T}$ and $I_{p^{\prime}}^{M}$, as well as the smooth term $\mathcal{L}_{Jdet}$.
%

The inference mode of NIR is much simpler: the pre-trained neural field takes the whole grid coordinates as input and outputs the deformations at all input coordinates. The warped volume $W$ is then obtained by sampling intensities from the moving volume $M$ given the deformed coordinates.

In Sec.~\ref{subsection:network_design}, we describe the network design of $\textit{NF}$. In Sec.~\ref{subsction:opt}, we go over several optimization components, including \textit{CS}, \textit{IS}, and the objective functions.  In Sec.~\ref{subsection:cascaded registration}, we present hybrid coordinate sampling scheme that strikes a balance between registration accuracy and regularity and maintain the optimization efficiency. 


\subsection{Network Design}
\label{subsection:network_design}
As illustrated in Fig.~\ref{fig:model_overview}, $\textit{NF}$ takes as input a 3D coordinate $\boldsymbol{p}\in\mathbb{R}^3$ in $T$ and outputs the corresponding coordinate $\boldsymbol{p}^{\prime}\in\mathbb{R}^3$ in $M$. The transformation from $\boldsymbol{p}$ to  $\boldsymbol{p}^{\prime}$ can be parameterized in two options: 1) use a neural field to directly predict the the displacement vector, or 2) use a neural field to predict the velocity vector, the integral of which leads to the deformation vector. 
Both neural displacement field and neural velocity field can be formulated as the Eq.~\ref{eq:neural_fields} and next we will look into the design of each component in our neural fields.

\subsubsection{Coordinate Encoding}
\label{subsubsection:pe}
Coordinate encoding module maps three-dimensional input coordinates to a higher-dimensional embedding \cite{mildenhall2020nerf,MatthewTancik2020FourierFL}. The mapping can be realized by a family of functionals $e_{i}: \mathbb{R}^{3} \rightarrow \mathbb{R}^{2}$, written as:

\begin{equation}
\label{eq:ce}
    \gamma(\boldsymbol{p})=\left[e_{1}(\boldsymbol{p}), e_{2}(\boldsymbol{p}), \ldots, e_{n}(\boldsymbol{p})\right]
\end{equation}

We follow the suggestion from \cite{MatthewTancik2020FourierFL}, encoding coordinates via Fourier mapping, such that 

\begin{equation}
\label{eq:ffn}
e_{i}(\boldsymbol{p})=\left[\cos \left(2 \pi \mathbf{\omega}_{i}^{\top} \boldsymbol{p}\right), \sin \left(2 \pi \mathbf{\omega}_{i}^{\top} \boldsymbol{p}\right)\right]^{\top},
\end{equation}

where $\mathbf{\omega}_i \in \mathbb{R}^{3}$ is randomly sampled i.i.d. from a Gaussian distribution with standard deviation $\sigma$. The higher the $\sigma$, the more likely the model will bias towards the high-frequency signal. In our experiments, $n$ and $\sigma$ are set to be 128 and 3 no matter which neural field and coordinate samplers we choose.

\subsubsection{Sinusoidal representation networks (SIRENs)}
On top of coordinate encoding layer, the main body of our neural field is a SIREN network \cite{sitzmann2020implicit}, in which all neurons are activated with sinusoidal functions, i.e., $\rho^{(\ell)} =sin$. Notably, the first layer of SIREN networks can be written as $\boldsymbol{z}^{(1)}=\sin \left(\omega_{0}\left(\boldsymbol{W}^{(0)} \boldsymbol{z}^{(0)}+\boldsymbol{b}^{(0)}\right)\right)$. Thus, similar to Fourier coordinate mapping,  SIRENs can also regulate the spectral bias of the network by adjusting the network hyperparameter $\omega_0$, which is set to be 30 for all our experiments.

\cite{yuce2022structured} reveals that the expressive power of coordinate-based MLP with sinusoidal encodings is equivalent to that of a structured signal dictionary, which is restricted to functions that can be expressed as a linear combination of certain harmonics of the coordinate encoding $\gamma(\boldsymbol{p})$. SIREN can be seen as the nested sinusoids and the few coefficients of this network are enough to represent signals with an exponentially large frequency support.  


\begin{figure}
    \centering
     \begin{subfigure}[b]{0.3\textwidth}
         \centering
         \includegraphics[height=0.5\textwidth]{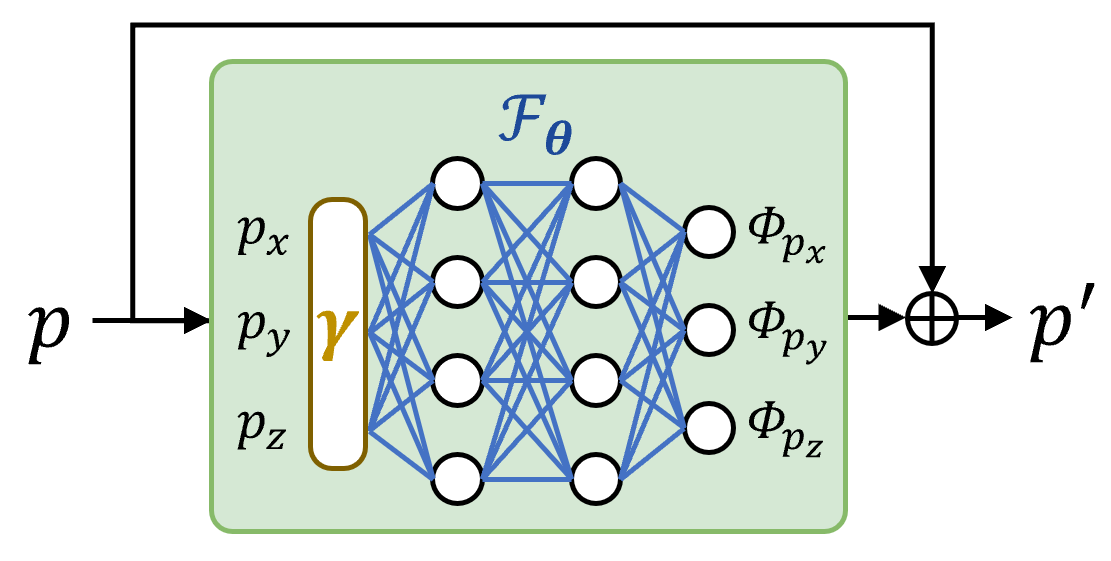}
         \caption{Neural Displacement Field}
         \label{fig:deformation_nf}
     \end{subfigure}
     \vfill
     \vspace{1em}
     \begin{subfigure}[b]{0.3\textwidth}
         \centering
         \includegraphics[height=0.5\textwidth]{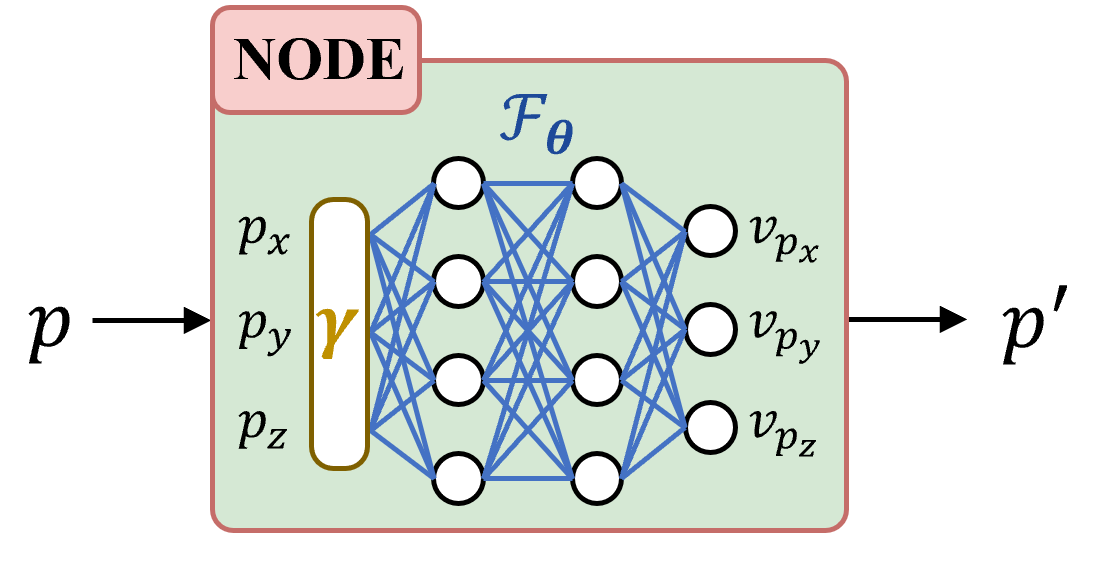}
         \caption{Neural Velocity Field}
         \label{fig:dd_nf}
     \end{subfigure}
\caption{\textbf{Neural Fields for Coordinate Deformations} -- In the above figure, blue modules indicate the parameters to be optimized. (a) illustrates the neural deformation field that directly transforms the coordinate $p$ in the target volume to the coordinate $p^{\prime}$ in the moving volume. (b) illustrate the neural velocity field which predicts the stationary velocity vector along the deformation trajectory from  $p$ to $p^{\prime}$. The neural velocity field plays as the dynamic function of a NODE solver and the final deformations are obtained via the integration of the predicted velocity vector.}
\label{fig:nfs}
\end{figure}

\subsubsection{Neural Displacement Field}
\label{subsubsubsection:ddf}
Neural displacement field $\mathcal{F}_{\theta}$ takes as input a 3D location $p$ in $T$ and outputs a displacement vector $\phi_p = [\phi_{p_x}, \phi_{p_y}, \phi_{p_z}]^T=\mathcal{F}_{\theta}(\boldsymbol{p})$. As a result, the deformed position $\boldsymbol{p}^{\prime}$ in $M$ is $\boldsymbol{p} + \phi_p$.

\subsubsection{Neural Velocity Field}
\label{subsubsubsection:dddf}
Under this option, our proposed framework can perform diffeomorphic image registration. Let $\Phi(\boldsymbol{p}, t): \Omega\subset\mathbb{R}^{3} \times [0, 1] \mapsto \Omega\subset\mathbb{R}^3$ define a continuous, invertible trajectory from the initial position $\boldsymbol{p}=\Phi(\boldsymbol{p}, 0)$ to the final position $\boldsymbol{p}^{\prime}=\Phi(\boldsymbol{p}, 1)$, satisfying such ordinary differential equation (ODE) and the initial condition:
\begin{equation}
    \frac{\partial \Phi(\boldsymbol{p}, t)}{\partial t}=\boldsymbol{v}(\Phi(\boldsymbol{p}, t), t) \quad \text { s.t. } \quad \Phi(\boldsymbol{p}, 0)=\boldsymbol{p},
    \label{eq.forward_ode}
\end{equation}
where $\boldsymbol{v}(\boldsymbol{p}, t): \Omega \times [0, 1] \mapsto \Omega$ indicates the velocity vector of coordinate $\boldsymbol{p}$ at time t. If $\boldsymbol{v}$ is Lipschitz continuous, a solution to Eq.~(\ref{eq.forward_ode}) exists and is unique in the interval $[0, 1]$, which ensures that any two deformation trajectories do not cross each other \cite{dupont2019augmented}. In this work, we assume that ${\boldsymbol{v}}$ is stationary and can be modeled via a neural field, written as $\mathcal{F}_{\theta}(\boldsymbol{p}) = [\boldsymbol{v}_{p_x}, \boldsymbol{v}_{p_y}, \boldsymbol{v}_{p_z}]^T$. 

The initial value problem (IVP) in Eq.~\ref{eq.forward_ode} can be solved with a Differentiable ODE Solver (NODE) \cite{chen2018neuralode} whose dynamic function is set to be $\mathcal{F}_{\theta}$. Considering the trade-offs between speed and accuracy, we choose the Fourth-order Runge-Kutta method (rk4) with step size of 0.25 as the ODE solver for our diffeomorohic registration experiments. In the forward pass, the deformed position $\boldsymbol{p}^{\prime}$ of position $\boldsymbol{p}$ can be estimated by integrating $\mathcal{F}_{\theta}(\boldsymbol{p})$ from $t=0$ to $t=1$ via NODE, formulated as

\begin{equation}
    \boldsymbol{p}^{\prime} = \Phi(\boldsymbol{p}, 1) = \Phi(\boldsymbol{p}, 0) + \int_{0}^{1} \mathcal{F}_{\theta}({\Phi(\boldsymbol{p}, t)}) \mathrm{d}t
    \label{eq.progressive_d_field}
\end{equation}

For backpropagation, NODE adopts the adjoint sensitivity method \cite{pontryagin1987mathematical}, which retrieves the gradient by solving the adjoint ODE backwards in time and allows solving with O(1) memory usage no matter how many steps the ODE solver takes.
\subsection{Optimization}
\label{subsction:opt}
In this section, we will introduce the intensity sampler, objective functions as well as coordinate sampler used in our NIR. 

\subsubsection{Intensity Sampler}
\label{subsubsection:intensity_sampling}
To utilize gradient-based optimization method, a differentiable intensity sampler is required to estimates the intensities of sub-voxel positions given source images. Same as \cite{jaderberg2015spatial,balakrishnan2019voxelmorph, mok2020fast, mok2020large}, we apply linear interpolation (other interpolation methods can also be applied) as intensity sampler, referred as $\mathcal{IS}^{\operatorname{linear}}$. Given a coordinate $\boldsymbol{c}$ and scans $S$, the intensity value at $\boldsymbol{c}$, referred to as $I_c^S$, is obtained based on the intensities of the eight surrounding voxels,
\begin{equation}
    I_c^S = \mathcal{IS}^{\operatorname{linear}}(\boldsymbol{c}, S) = \sum_{\boldsymbol{c}_i \in \mathcal{Z}\left(\boldsymbol{c}\right)} S(\boldsymbol{c}_i) \prod_{d \in\{x, y, z\}}\left(1-\left|\boldsymbol{c}_{i, d}-\boldsymbol{c}_{d}\right|\right),
\end{equation}
where $\mathcal{Z}\left(\boldsymbol{c}^{\prime}\right)$ denotes the voxel neighbors of $\boldsymbol{c}$, $d$ iterates the dimension index, and $S(\boldsymbol{c_i})$ indicates the intensity value at voxel $c_i$ on volume S.

\subsubsection{Objective Functions}
\label{subsubsection:loss}
Local normalized cross-correlation is adopted to measure the intensity similarity. Let $\bar{I_c^S}$ denote the intensity mean of local region centering at position c on volume $S$. In our experiments, $\bar{I_c^S}=\frac{\sum_{c_{i}}I_{c_i}^S}{w^3}$, where $\boldsymbol{c}_i$ iterates over the local region in the size of $w^3$ with $w=9$. Then local normalized corss-correlation can be defined as below:
\begin{align}
  {\rm LNCC}(T, M, \boldsymbol{p}, \boldsymbol{p^{\prime}}) = 
  \frac{\left[\sum_{p_{i},p_{i}^{\prime}}(I_{p_i}^T-\bar{I_p^T})(I_{p_i^{\prime}}^M-\bar{I_{p^{\prime}}^M})\right]^2}{\left[\sum_{p_{i}}(I_{p_i}^T-\bar{I_p^T})^2\right]\left[\sum_{p_{i}^{\prime}}(I_{p_{i}^{\prime}}^M-\bar{I_{p^{\prime}}^M})^2\right]},
\label{eq:lncc}
\end{align}
where $\boldsymbol{p}$ denotes the sampled position in the coordinate of target volume $T$, and $\boldsymbol{p}^{\prime}$ denotes deformed position in the coordinate of moving volume $M$.

As for the regularization term, we follow \cite{mok2020fast} to impose the Jacobian determinant penalty on the predicted deformation field. The Jacobian matrix of the deformation field $\phi$ at a position $\boldsymbol{p}$, notated as $J_{\phi}(\boldsymbol{p})$, is given by:

\begin{equation}
    J_{\phi}(\boldsymbol{p})=\begin{bmatrix}
\frac{\partial \phi_{p_x}}{\partial x} & \frac{\partial \phi_{p_x}}{\partial y} & \frac{\partial \phi_{p_x}}{\partial z} \\
\frac{\partial \phi_{p_y}}{\partial x} & \frac{\partial \phi_{p_y}}{\partial y} & \frac{\partial \phi_{p_y}}{\partial z} \\
\frac{\partial \phi_{p_z}}{\partial x} & \frac{\partial \phi_{p_z}}{\partial y} & \frac{\partial \phi_{p_z}}{\partial z},
\end{bmatrix}
\label{eq:jac}
\end{equation}

where $\phi_p = [\phi_{p_x}, \phi_{p_y}, \phi_{p_z}] \in \mathbb{R}^{3}$ denotes the deformation vector at position $\boldsymbol{p}$. 
If $|J_{\phi}(\boldsymbol{p})|$ is positive, it is suggested that the deformation field preserves the local orientation near $\boldsymbol{p}$. Conversely, if $|J_{\phi}(\boldsymbol{p})|$ is negative, the deformation field reverses the local orientation around $\boldsymbol{p}$. Thus, the local orientation consistency constraint can be defined as
\begin{equation}
   {\rm LOCC}(\boldsymbol{p})=\max(0, -|J_{\phi}(\boldsymbol{p})|), 
\end{equation}
which only penalizes the regions with negative Jacobian determinants. In our experiment, $J_{\phi}(\boldsymbol{c})$ is approximated as the differences between neighboring deformation vectors.   

Overall, the objective function $\mathcal{L}$ can be expressed as the weighted sum of the intensity similarity $\mathcal{L}_{sim}$ and the Jacobian determinant regularization $\mathcal{L}_{reg}$, where
\begin{align}
    \mathcal{L}_{sim} &= \frac{1}{N}\sum_j-{\rm LNCC}(T, M, \boldsymbol{p}_j) \\
    \mathcal{L}_{reg} &= \frac{1}{N}\sum_j{\rm LOCC}(\boldsymbol{p}_j) \\
    \mathcal{L} &= \mathcal{L}_{sim} + \lambda_{reg} \cdot \mathcal{L}_{reg}
\label{eq:loss}
\end{align}
Here, $N$ is the toal number of sampled locations per optimization iteration and $\boldsymbol{p}_j$ denotes the $j$th location sampled in one batch.

\begin{figure*}
    \hfill
    \begin{subfigure}[t]{0.15\textwidth}
        \centering
        \includegraphics[height=0.8\textwidth,keepaspectratio,valign=t]{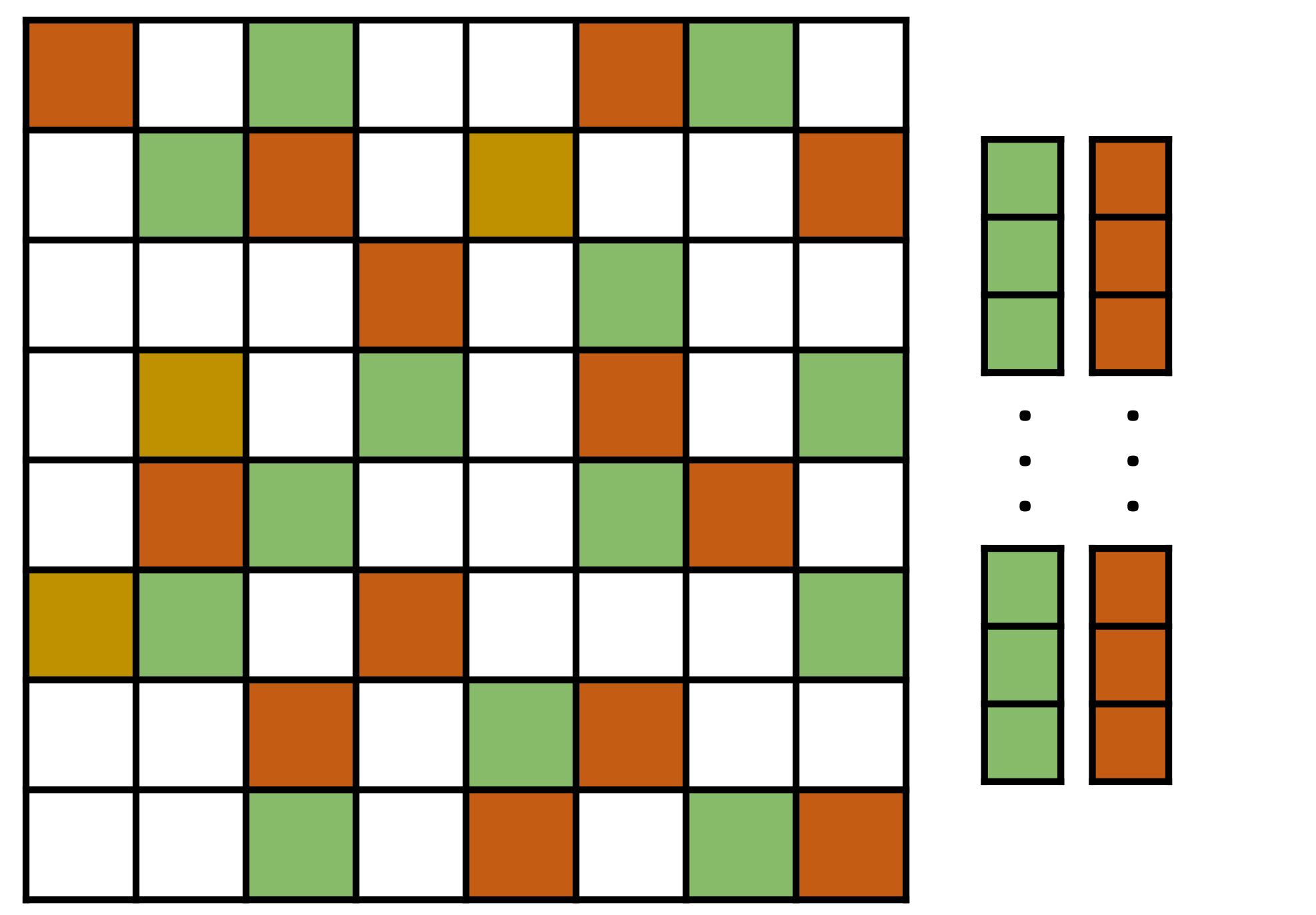}
        \caption{Random Sampler}
        \label{fig:random_cs}
    \end{subfigure}
    \hfill
    \begin{subfigure}[t]{0.15\textwidth}
        \centering
        \includegraphics[height=0.8\textwidth,keepaspectratio,valign=t]{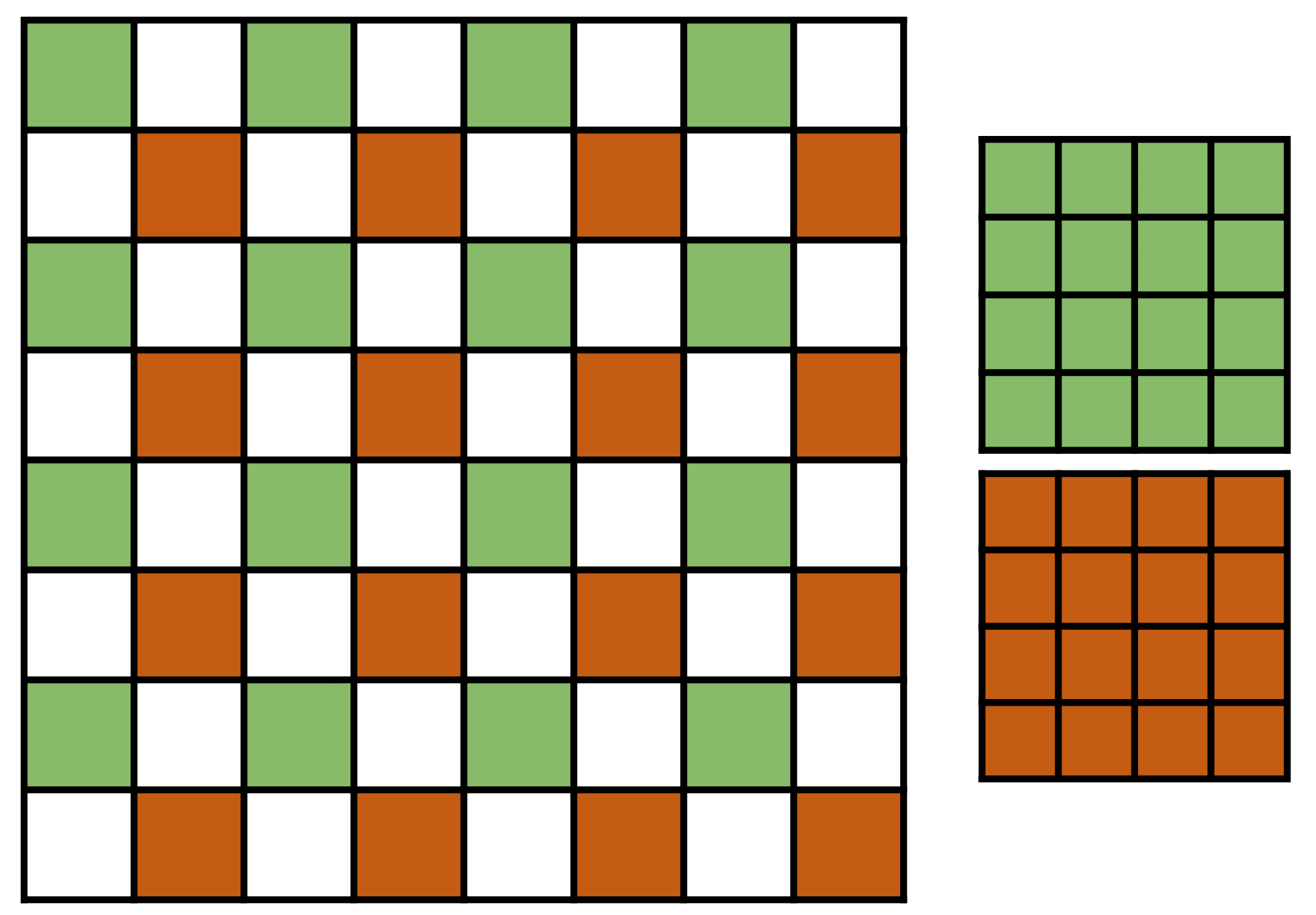}
        \caption{Downsize Sampler}
        \label{fig:downsize_cs}
    \end{subfigure}
    \hfill
    \begin{subfigure}[t]{0.15\textwidth}
        \centering
        \includegraphics[height=0.8\textwidth,keepaspectratio,valign=t]{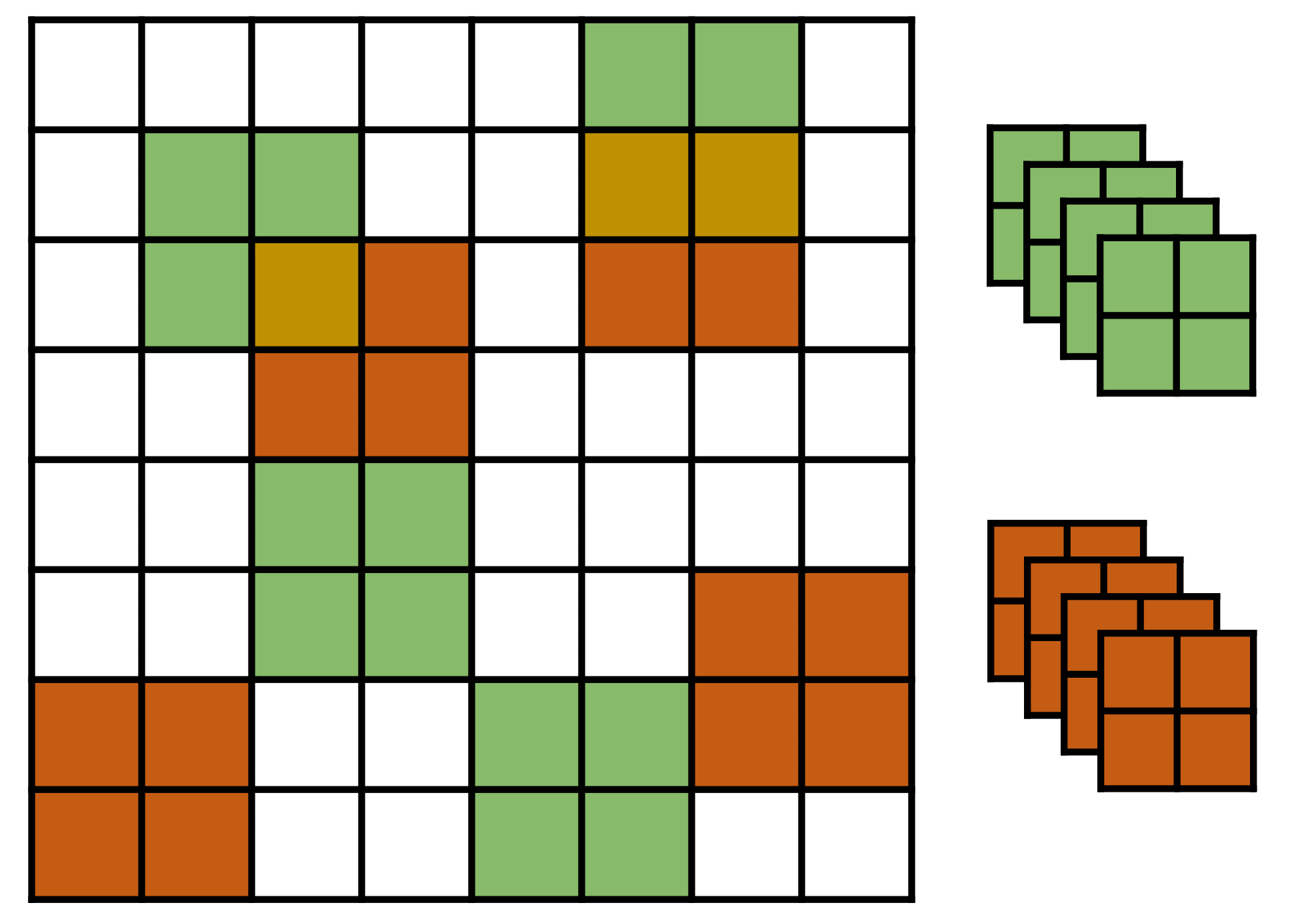}
        \caption{Mini-patch Sampler}
        \label{fig:minipatch_cs}
    \end{subfigure}
    \hfill
    \begin{subfigure}[t]{0.25\textwidth}
        \centering
        \includegraphics[height=0.8\textwidth,keepaspectratio,valign=t]{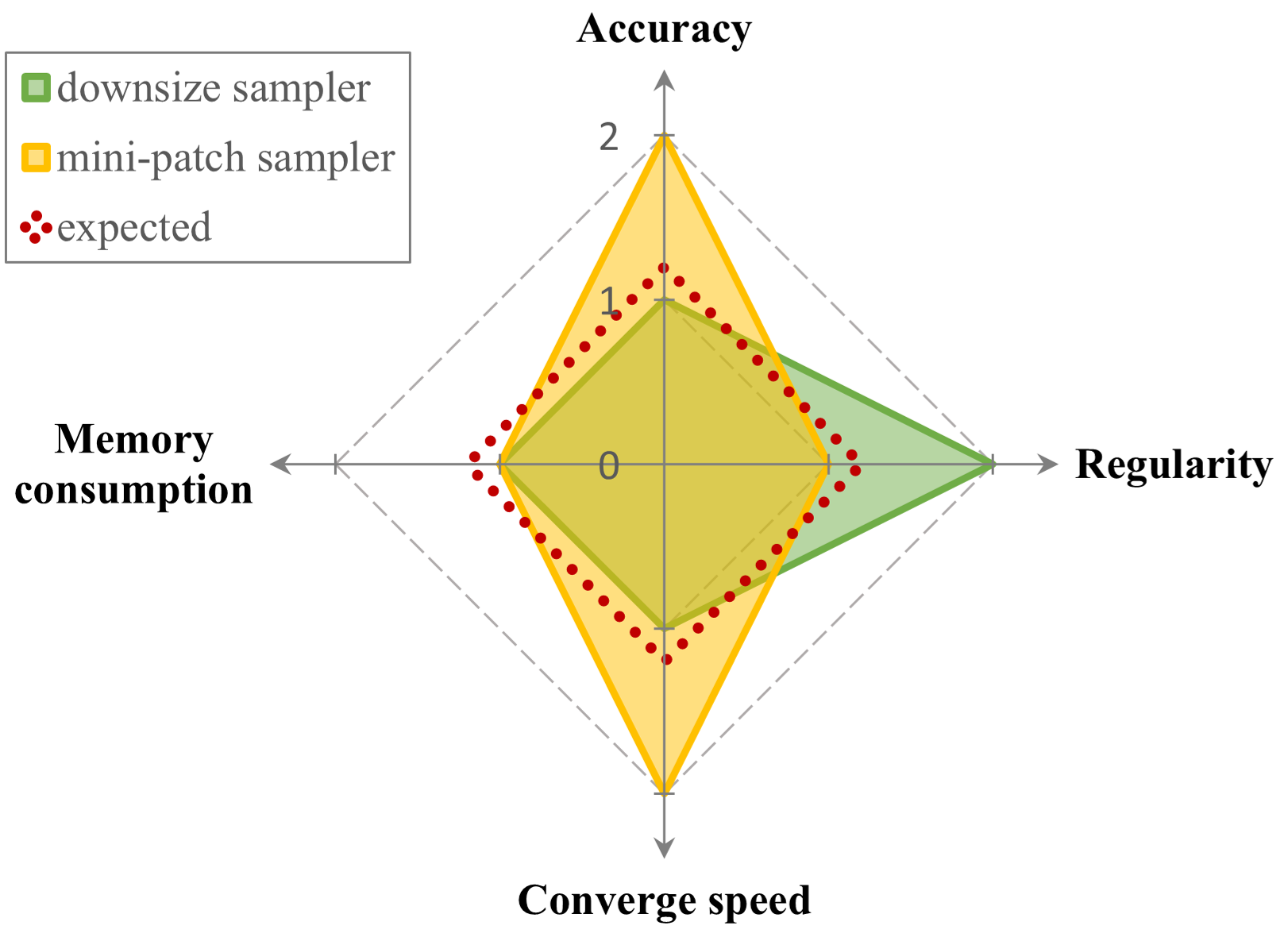}
        \caption{Performance Comparison}
        \label{fig:cs_performance_comparison}
    \end{subfigure}
    \hfill
    \vspace{-2.5em}
    \vfill
    \hspace{7em}
    \begin{subfigure}[t]{0.45\textwidth}
        \centering
        \includegraphics[width=\textwidth]{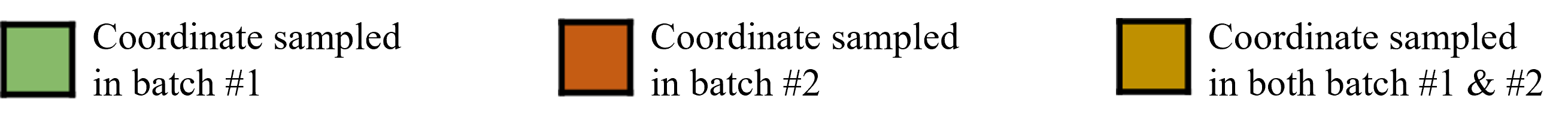}
        \label{fig:cs_notion}
    \end{subfigure}
    \hfill
\vspace{-1em}
\caption{\textbf{Coordinate Samplers and Performance Comparisons}. (a), (b) and (c) illustrate sampling 16 coordinates per batch from total 64 2D coordinates with three kinds of coordinate samplers. (d) ranks the registration performance of NIR models optimized with two practical coordinate samplers (downsize sampler and mini-patch sampler) in four aspects. The higher ranking in each dimension indicates better performance in that aspect. As is shown in (d), consuming almost the same GPU memory during optimization, compared to NIR optimized with the mini-patch sampler, NIR optimized with the downsize sampler can take less time to converge to a more accurate registration results with more violations in topology preserving. The expected solution, as indicated by the red-dot line, should be of great performance in both registration accuracy and regularity with no or modestly extra computations. For the numerical results supporting the ranking in plot (d), please refer to Tab.~\ref{tab:comp_among_cs}.}
\label{fig:cs}
\end{figure*}

\subsubsection{Coordinate Sampler}
\label{subsubsection:coords_sampling}
To optimize the parameters of our neural fields, we apply mini-batch stochastic gradient descent method. In other words, we sample a subset of coordinates of the whole image grid to update the model parameters per iteration in optimization. Next, we will discuss three different coordinate samplers: random sampler, downsize sampler and mini-patch sampler. 

\textit{Random Sampler} (Fig.~\ref{fig:random_cs}) is most commonly used in coordinate-based neural networks \cite{sitzmann2020implicit, niemeyer2019occupancy, chen2021learning} because the coordinates sampled via a random sampler are distributed across the whole grid and the unbiased sampled coordinates allow for the more stable optimization. But random coordinate sampler is inapplicable in our case. To compute $LNCC$, we need to search closest coordinates among all sampled coordinates to estimate the local intensity mean and correlation, whose consequence is that the optimization speed can be significantly impeded by the searching time. Moreover, randomly sampling coordinate will bring about larger memory consumption for calculating $LOCC$. As we mentioned in Sec.~\ref{subsubsection:loss}, we approximate the Jacobian matrix of the deformation field by discretizing the image coordinate space, asking for the coordinates to be sampled in a spatial regularity. If the sampled coordinates are distributed randomly, the Jacobian matrix requires extra memory for the second-order derivatives of deformation field with respect to model parameters during optimization. After all, considering the time and memory deficiency, random coordinate sampler is an impractical choice for our NIR. 

\textit{Downsize Sampler} samples coordinates with specific step size in each dimension as shown in Fig.~\ref{fig:downsize_cs}. Coordinates sampled by downsize sampler can well cover the entire image coordinate space but the approximation of Jacobian matrix might be of more flaws due to downsizing. The consequence is, the neural fields optimized via downsize coordinate sampler achieve great alignment accuracy but relatively bad local orientation consistency in deformations. The down-sampling step size used in downsize coordinate sampler is set as 3 along all dimensions in our experiments.  


\textit{Mini-Patch Sampler} randomly selects multiple high-resolution small coordinate blocks as shown in Fig.~\ref{fig:minipatch_cs}. Compared to downsize coordinate sampler, it can provide more accurate Jacobian matrix approximation but the drawback lies in the computation of local normalized cross-correlation. Specifically, the extensive padding operations along the patch borders result in the inaccurate local normalized cross-correlation. Thus, the neural fields optimized via mini-patch coordinate sampler are good at registration regularity but bad at alignment accuracy. In our experiments, the mini-patch coordinate samplers randomly select 5 patches in size of $32\times32\times32$ per optimization iteration.

Fig.~\ref{fig:cs_performance_comparison} demonstrates the rank of the registration performance of two candidate coordinate samplers with the spatial regularity in four criteria --- accuracy, regularity, memory consumption, and converge speed. It is apparent in Fig.~\ref{fig:cs_performance_comparison} that no coordinate sampling strategy can outperform the others in all criteria. Downsize coordinate sampler is good at criteria in all aspects but the registration regularity, which happens to be the strength of mini-patch coordinate sampler. The expected solution should have high registration accuracy, minor distortions in the deformation field, rapid converge rate as well as little memory consumption, as indicated by the red-dot region in Fig.~\ref{fig:cs_performance_comparison}. 
Please refer to Sec.~\ref{subsec:metrics} and Sec.~\ref{subsubsection:coords_sampling} for more details in the evaluation metrics and quantitative comparisons of downsize sampler and mini-patch sampler.

\subsection{NIR with Hybrid Coordinate Sampler}
\label{subsection:cascaded registration}

\subsubsection{Overview}
\label{subsubsection:hybrid_mirnf_overview}
\begin{figure}
     \centering
     \begin{subfigure}[b]{0.45\textwidth}
         \centering
         \includegraphics[height=0.5\textwidth]{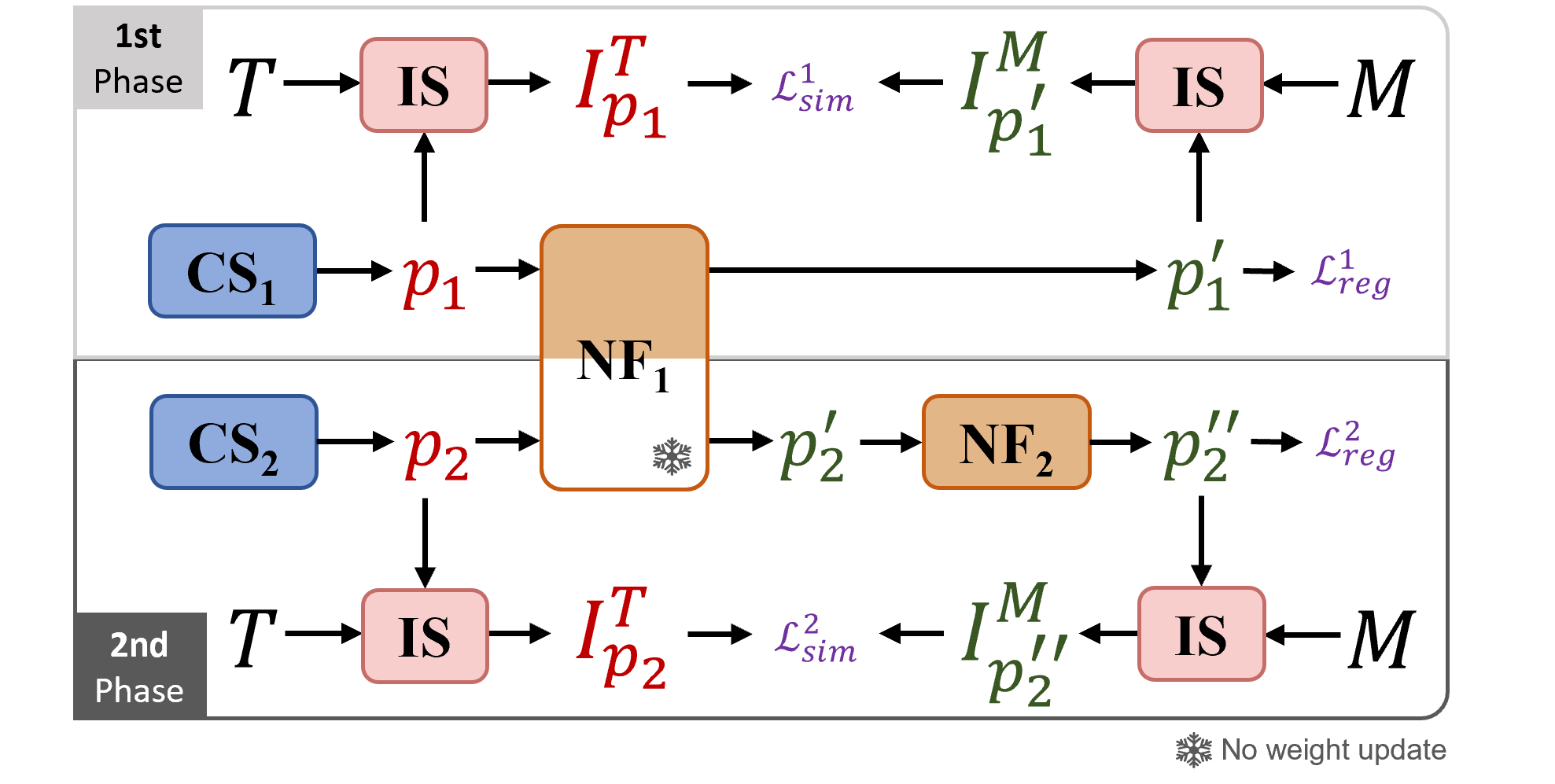}
         \caption{Optimization}
         \label{fig:hybrid-framework-opt}
     \end{subfigure}
     \hfill
     \vfill
     \begin{subfigure}[b]{0.45\textwidth}
         \centering
         \includegraphics[height=0.35\textwidth]{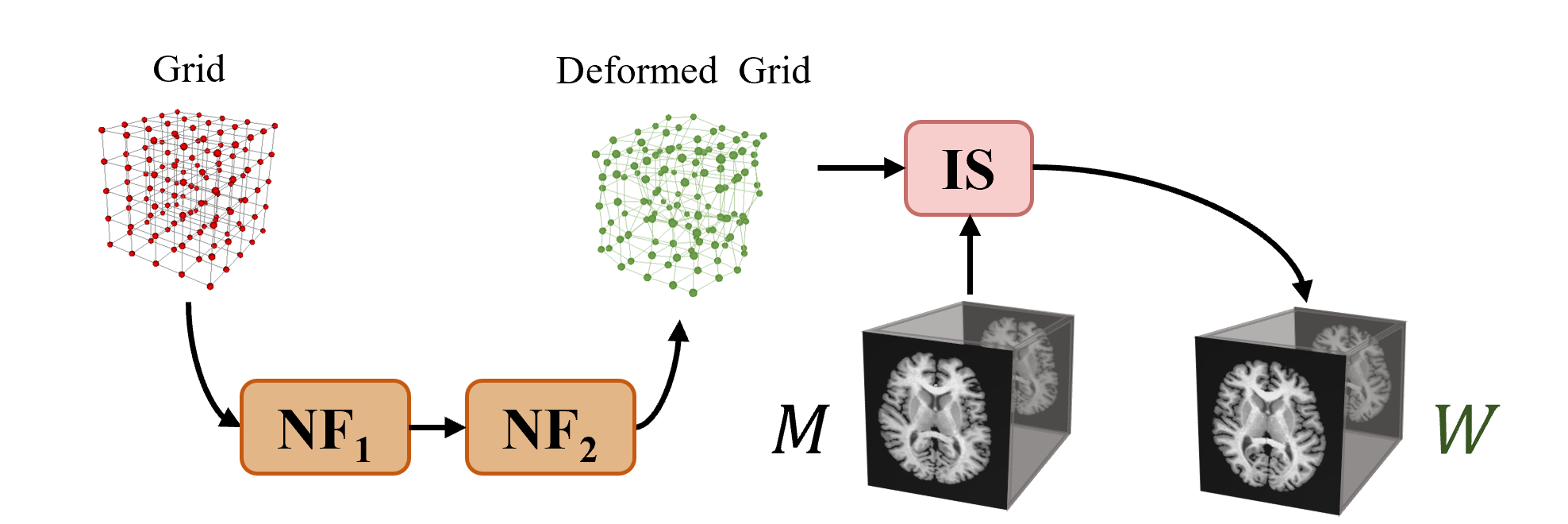}
         \caption{Inference}
         \label{fig:hybrid-framework-inf}
     \end{subfigure}
\caption{\textbf{Overview of NIR with Hybrid Coordinate Sampling Scheme}. The optimization is composed of two phases, in which two neural fields ($NF_1$ and $NF_2$) are optimized separately. In the first phase, $NF_1$ is optimized with the downsize coordinate sampler ($CS_1$) for 200 iterations. In the second phase, with the mini-patch coordinate sampler ($CS_2$), fixed $NF_1$ provides the initial deformations and only $NF_2$ is optimized. During inference, NIR with hybrid coordinate sampler requires grid coordinates to pass through two neural fields in sequence to get the deformed coordinates.}
\label{fig:hybrid_model_overview}
\end{figure}

We intend to enhance the complementarity of the downsize and mini-patch coordinate samplers without the substantial increase in memory and time consumption during optimization. To this end, we propose a hybrid coordinate sampler which performs two different coordinate sampling strategies in two phases of optimization. As shown in Fig.~\ref{fig:hybrid_model_overview}, NIR with a hybrid coordinate sampler consists of two concatenated neural fields optimized separately.
The first neural field ($NF_1$) takes in charge of the rough alignment between the moving and target scans and the residual transformation is completed by another neural field ($NF_2$).
In inference, $NF_1$ and $NF_2$ deform the whole grid in cascade, which means the output of $NF_1$ is taken as the input of $NF_2$ and then $NF_2$ outputs the final deformed grid. As for optimization, the parameters of $NF_1$ and $NF_2$ are updated with the downsize sampler $CS_1$ and mini-patch sampler $CS_2$ separately in two phases as depicted in Fig.~\ref{fig:hybrid-framework-opt}. 

\subsubsection{Optimization}
\begin{figure*}[!ht]
     \centering
     \includegraphics[width=0.8\textwidth]{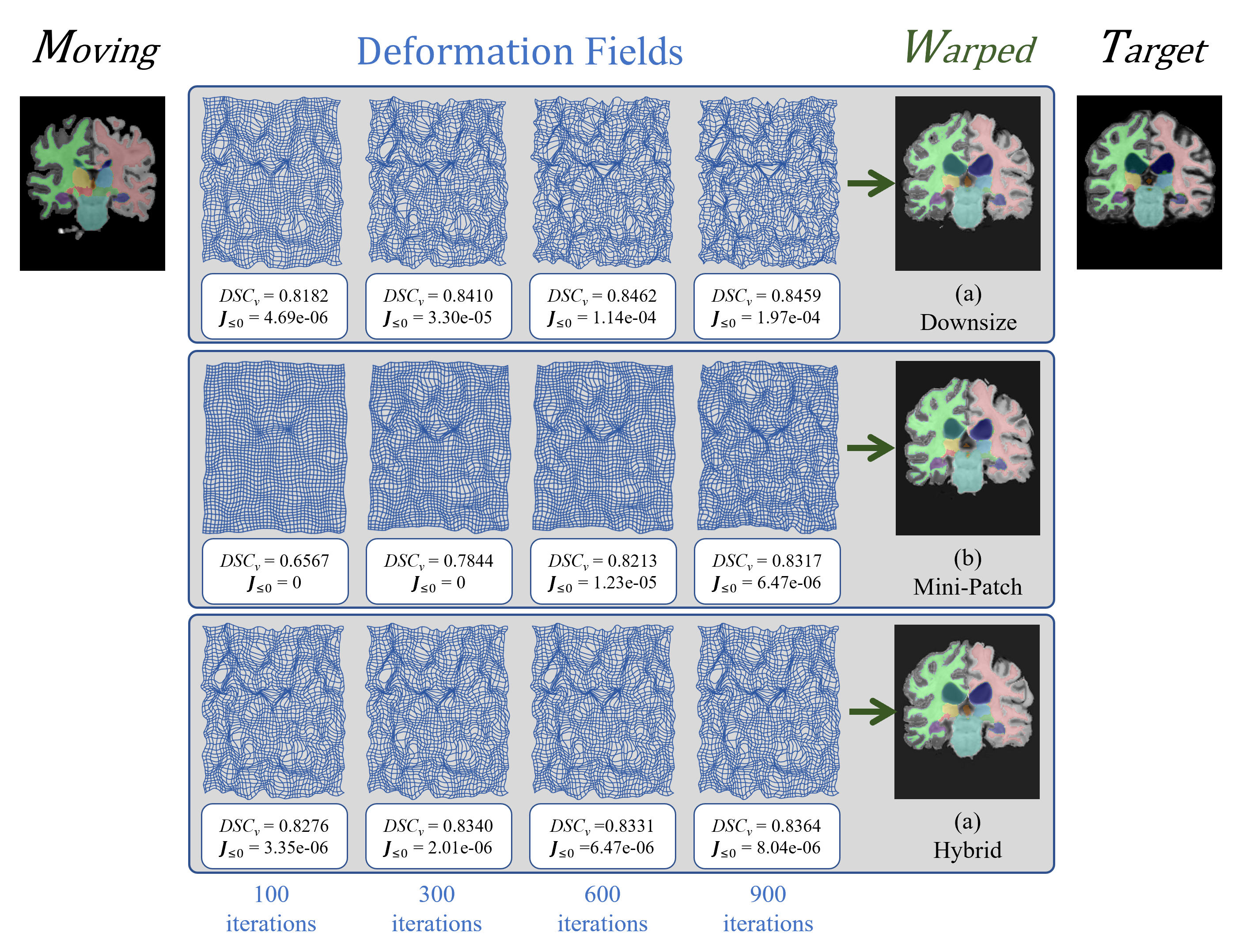}
\caption{\textbf{Comparison of different coordinate samplers} (a), (b) and (c) are registration results of NIR optimized with the downsize sampler, mini-patch sampler and hybrid NIR. The above image pair are 'OASIS\_OAS1\_0001\_MR1' ($T$) and 'OASIS\_OAS1\_0002\_MR1' ($M$) from the OASIS dataset and we present the registration results over the optimization iterations, generated by the differomorphic NIR. DSC and $J_{\leq 0}$ are the evaluation metrics for registration accuracy and regularity separately. Details about the dataset and evaluation metrics can be found in Sec.~\ref{sec:exp}. } 
\label{fig:hybrid_framework_explaination}
\end{figure*}

The example in Fig.~\ref{fig:hybrid_framework_explaination} demonstrates that, the downsize sampler can generate more accurate registration in the price of more distortions in the deformation field while the mini-patch sampler tends to provide the over-smooth deformation fields and results in much slower convergence speed. We also noticed that, in the very early stage of optimization, the regularity of deformation field from NIR optimized with the downsize sampler is well-preserved and at the same time, registration accuracy is quite decent. But as optimization time grows, the fraction of positions with a negative Jacobian determinant increases a lot. 
One possible explanation is that, owing to the spectral bias of neural networks, the fields tends to reconstruct the lower-frequency signal in the beginning of optimization. 

The optimization strategy for NIR with a hybrid sampler is motivated by the above observations and is conducted in two phases.
In the first phase, $NF_1$ is optimized with the downsize coordinate sampler $CS_1$ for a short time. After the first-phase optimization, $NF_1$ is able to generate the smooth and relatively accurate transformation. Then the goal of the second-phase optimization is to let $NF_2$ complete the transformation left unfinished by $NF_1$. We prefer a neural registration field that can align the more detailed structures and doesn't mess up the underlying topology in the second phase of optimization. For this reason, the input coordinate $p_2$ for the second-phase optimization are sampled by the mini-patch sampler $CS_2$ and the initial deformed positions $p_2^{\prime\prime}$ for $NF_2$ are predicted by $NF_1$. Notably, taking optimization stability and memory efficiency into account, only $NF_2$ is optimized in the second-phrase optimization while weights of $NF_1$ are frozen. 

Similar to NIR, the optimization objective functions of NIR with a hybrid sampler is given by:

\begin{equation}
    \mathcal{L} = w_1 \cdot (\mathcal{L}_{sim}^{1} + \lambda_{reg} \cdot \mathcal{L}_{reg}^{1}) + w_2 \cdot (\mathcal{L}_{sim}^{2} + \lambda_{reg} \cdot \mathcal{L}_{reg}^{2}),
    \label{eq:hybrid_opt}
\end{equation}

where $w_1 = 1, w_2 = 0$ in the first phase of optimization and $w_1 = 0, w_2 = 1$ in the second phase. $\mathcal{L}_{sim}^1$, $\mathcal{L}_{sim}^2$, $\mathcal{L}_{reg}^1$ and $\mathcal{L}_{reg}^2$ follows the same definition as introduced in Sec.~\ref{subsubsection:loss}. 

\section{Experiments}
\label{sec:exp}

\subsection{Dataset}
\label{subsec:data}
All our experiments are conducted on two public 3D brain MR datasets -- Mindboggle101 and OASIS.

\textbf{Mindboggle101} \cite{klein2012101} consists of 101 T1-weighted MR scans of healthy participants coming from 5 data sources. We select 31 cortical regions as \cite{xu2019deepatlas} for evaluation. 

\textbf{OASIS} dataset contains 416 T1-weighted MR images of subjects aged from 18 to 96 including individuals with early-stage Alzheimer’s Disease (AD). \cite{hoopes2021hypermorph} annotated 35 anatomical structures associated with OASIS dataset and 27 of them are selected for the performance evaluation in experiments.

All MRI images used in our experiments are pre-processed by the same procedures, which sequentially are skull stripping, resampling to $1mm\times1mm\times1mm$ spacings, affinely align to MNI template of T1-weighted MRI imaging \cite{fonov2009unbiased, fonov2011unbiased} and cropping to size of $160\times192\times144$. Since all images have already been aligned to the MNI template, we focus on the non-linear deformation between pair of images in our experiments.
\subsection{Implementation Details}
\label{subsction:imple_details}
The neural displacement field in Sec.\ref{subsubsubsection:ddf} contains 4 fully connected layers with hidden feature in the length of 256.
Due to the computation inefficiency of NODE, we design shallower SIREN models for neural velocity field in Sec,\ref{subsubsubsection:dddf}, which contains 3 fully connected layers with 256 hidden feature size.
The network parameters are updated using Adam optimizer \cite{kingma2014adam} with a learning rate of $1e^{-4}$. During the optimization, ${\lambda_{reg}}$ are set as 1000 for our displacement-based deformable registration methods and 100 for our diffeomorphic registration methods. For NIR, the maximum optimization iterations is set to be 900. For NIR with a hybrid coordinate sampler, the first phase is optimized for 200 iterations and the second phase is further optimized for 900 iterations. 
Our framework is implemented with PyTorch \cite{paszke2019pytorch} and all experiments are deployed on a machine with a NVIDIA GTX 2080Ti GPU and an Intel i7-7700K CPU.
\subsection{Experimental Setup}
\label{subsec:exp_setup}

All medical image registration methods in our experiments aim to transform a moving volume to a target volume. If the structure labels in association with the moving volume are available, we can map the structures onto the target volume via the transformation obtained from the image registration task. The registration results are evaluated on the similarity between target volumes (structures) and warped volumes (structures), and the local orientation consistency of the deformation fields.

In this paper, we conduct three groups of unsupervised registration experiments where (1) 5 moving scans and 40 target scans both come from the Mindboggle101 dataset; (2) 5 moving scans and 40 target scans both come from the OASIS dataset; (3) 3 moving scans of healthy brains come from the Mindboggle101 dataset and 20 target scans with Alzheimer's disease come from the OASIS dataset. In experiment (1), 45 image scans with structure labels are randomly selected for test and the rest data in the Mindboggle101 dataset are used for training the learning-based models in comparisons. In experiment (2),
we randomly select 45 cases for test and 250 of the rest data are randomly selected as the training data for the learning-based methods in comparisons. The purpose of the first two experiments is to assess the applicability of our proposed methods in the brain MRI registration task. In experiment (3), 3 moving volumes are randomly selected from the Mindboggle101 dataset and 20 target volumes are randomly selected among the cases with the Clinical Dementia Rating (CDR) larger than 0.5 from the OASIS dataset, i.e., patients who have been diagnosed with moderate Alzheimer's disease. By comparing with the learning-based methods, this experiment can help reveal the performance robustness of our proposed optimization-based methods.

In order to determine some hyper parameters of our proposed methods, such as the learning rate and the regularization weight, 15 image pairs are randomly selected from the Mindboggle101 dataset to be the validation set. These hyper parameters are fixed across the above three groups of experiments.

\subsection{Methods in comparisons}
\label{subsec:baseline}

In Sec.~\ref{subsubsection:neural_field}, we have introduced two types of neural fields for the displacement-based deformable registration and diffeomorphic registration respectively, both of which can be integrated into the framework of NIR (Sec.~\ref{subsection:network_design}) and NIR with a hybrid coordinate sampler (Sec.~\ref{subsection:cascaded registration}). We provide several options of running NIR and their names are listed in Tab.~\ref{tab:methods}.
Next, we will go through the baseline methods and their training or optimization recipes in our experiments.


\begin{table}[]
\centering
\caption{
\textbf{Names of Options under NIR Framework}
}
\begin{tabular}{lcc}
\toprule
Name   & \begin{tabular}[c]{@{}c@{}}Coordinate \\ Sampler\end{tabular} & \begin{tabular}[c]{@{}c@{}}Neural Field \\ Type\end{tabular} \\ \midrule
NIR-D      & Downsize         & Displacement              \\
NIR-H      & Hybrid           & Displacement              \\
NIR-D-Diff & Downsize         & Velocity                  \\
NIR-P-Diff & Mini-patch       & Velocity                  \\
NIR-H-Diff & Hybrid           & Velocity                  \\ \bottomrule
\end{tabular}
\label{tab:methods}
\end{table}


VoxelMorph is a well-known learning-based method enabling the pairwise, deformable 3D medical image registration. Instead of optimizing the objective function for each pair of images which might be time-consuming, it directly predicts the desired deformation by learning from the large training dataset. In our experiments, we followed the original setting and trained the Voxelmorph using $LNCC$ as the similarity metric with a regularization term weighted by 1 to enforce the smoothness of the predicted displacement field.

SYM\_Net is the state-of-the-art learning-based method which provides the diffeomorphic deformable image registration. Unlike VoxelMorph which directly predict the displacement field, SYM\_Net is a symmetric image registration method which maximize the image similarity inside the space of diffeomorphic deformation and estimate the forward and backward transformation simultaneously. We trained the SYM\_Net using $LNCC$ with the parameters suggested by their original paper. Specifically, the weights for penalizing the negative jacobian determinant, enforcing the smoothness of velocity field and constraining the bias for bidirectional velocity field are set to 1000, 3 and 0.1 respectively.

SyN is an popular diffeomorphic registration method and we applied the implementation by DIPY \cite{garyfallidis2014dipy}. We take CC as the metric with the sampling radius as 4 and the standard deviation of the Gaussian smoothing kernel as 2. The results reported in Tab.~\ref{tab:results_comp} are achieved with the maximum optimization iterations as \{100, 50, 25\} at each level. We didn't used their official implementation in the Advanced Normalization Tools (ANTs) \cite{avants2009advanced} because it will take over one hour to register one pair of images with CC as the metric with only \{40, 20\} optimization iterations at two scales, which is unacceptable.

NiftyReg is the fast free-form deformation algorithm for non-rigid registration. In our experiments, cubic B-spline interpolation is used to deform moving volumes to optimize $LNCC$ image similarity with the squared Jacobian determinant log as a penalty term. The standard deviation of the Gaussian kernel and the weight of the penalty term are set to be 40 and 0.01 separately, as suggested by \cite{shen2019networks}. In addition, three scales are used in optimization with the maximum optimization iterations as \{1200, 600, 300\} for each scale.
 
Grid is our implementation of a Demons-based deformable image registration method \cite{pennec1999understanding} optimized by gradient decent, similar to what Autograd Image Registration Laboratory (airlab)\cite{sandkuhler2018airlab} did. Grid applies the same optimization object functions, intensity sampler and optimizer as NIR-D. However, instead of modeling the deformation field as a coordinate-based MLP, Grid directly optimizes the displacement vector of each grid coordinate via gradient decent. Also, the whole grid coordinates are sampled in each optimization iteration. We don't intend to fully reproduce the original Demons via PyTorch but the expressive power of neural fields in modeling deformation fields can be justified if NIR-D outperforms Grid significantly.

\subsection{Evaluation Metrics}
\label{subsec:metrics}

All methods in comparisons aim to map the moving volumes to the target volumes. If the moving volume and target volume come from the same dataset, the Dice's coefficient ($\displaystyle DSC$) and the ratio of coordinates with non-positive Jacobian determinant ($\mathbf{J}_{\leq 0}$) are used for evaluation. If the moving volume and target volume don't share the same annotations, we will apply Structural Similarity Index ($\displaystyle SSIM$) and $\mathbf{J}_{\leq 0}$ for evaluation.

Two types of $\displaystyle DSC$ -- volumetric $\displaystyle DSC$ for OASIS data and surface $\displaystyle DSC$ for Mindboggle101 data, are used to evaluate the overlap between two regions. Given the target mask $\mathcal{M}_T$ and warped mask $\mathcal{M}_W$, the volumetric $\displaystyle DSC$ for  can be written as

\begin{equation}
    \displaystyle DSC_v=\frac{2\left|\mathcal{M}_T \cap \mathcal{M}_W\right|}{\left|\mathcal{M}_T\right|+\left|\mathcal{M}_W\right|}
\label{eq:dsc_v}
\end{equation}

As pointed out by \cite{reinke2021common}, the volumetric measurements may lead to the same evaluation score given substantially different regions with complex shapes. In such cases as cortical structures, boundary-based measures are preferred. To be more specific, we use the surface $\displaystyle DSC$ introduced by \cite{nikolov2018deep} to evaluate the alignment accuracy in experiment (1). The surface $\displaystyle DSC$ assesses rather than the overlap of two volumetric regions but two surfaces within a specific tolerance $\tau$, formulated as

\begin{equation}
    \displaystyle DSC^{(\tau)}_s=\frac{\left|\mathcal{S}_{T} \cap \mathcal{B}_{W}^{(\tau)}\right|+\left|\mathcal{S}_{W} \cap \mathcal{B}_{T}^{(\tau)}\right|}{\left|\mathcal{S}_{T}\right|+\left|\mathcal{S}_{W}\right|},
\label{eq:dsc_s}
\end{equation}

where $\mathcal{S}_{i}$ refers to the surfaces of mask $\mathcal{M}_i$, and $\mathcal{B}_{i}^{\tau}$ denotes the border regions for the surface $\mathcal{S}_{i}$ within a tolerance $\tau$, which is 1mm in our experiments. For more details about the surface $\displaystyle DSC$, please look into \cite{nikolov2018deep}. Both volumetric and surface $\displaystyle DSC$ range from 0 to 1 and higher score represents better registration accuracy. The final reported scores are the average $\displaystyle DSC$ of all structures over all pairs.

$\mathbf{J}_{\leq 0}$ measures the regularity of deformation fields as the ratio of coordinates with non-positive Jacobian determinant. The Jacobian matrix represents the derivatives of the deformations, indicating the property of the local deformation field. Only the local regions with positive Jacobian determinant is transformed with topology-preservation and invertibility, so larger $\mathbf{J}_{\leq 0}$ signifies worse registration regularity. The calculation of the Jacobian matrix of deformations is given in Sec.~\ref{subsubsection:loss}.

$\displaystyle SSIM$ \cite{wang2004image} is a weighted sum of three comparison measurements between two images: luminance, contrast and structure. $\displaystyle SSIM$ ranges between 0 and 1, with larger values representing higher similarity between image pairs. Please refer to the original paper \cite{wang2004image} for the calculation details.

\subsection{Quantitative Comparisons with Baselines}
\label{subsec:results_comp}
\begin{table*}[]
\centering
\caption{
\textbf{Registration Performance Comparison on Mindboggle101 dataset and OASIS dataset.} The GPU memory consumption for the learning-based methods are "training consumption | inference consumption", but for our proposed methods, are just maximum memory consumption during optimization. Surface Dice's Cofficient within 1mm tolerance ($\displaystyle DSC_s^{1mm}$) and Volumetric Dice's Cofficient ($\displaystyle DSC_v$) are respectively applied in Mindboggle101 dataset and OASIS dataset for registration accuracy. The ratio of coordinates with non-positive Jacobian determinant ($\mathbf{J}_{\leq 0}$) are used to evaluate the registration regularity. The GPU memory consumption in optimizing the hybrid MINRF models varies in two phases because the numbers of sampled coordinates in two phases are different. Specifically, the number of coordinates sampled by the downsize sampler and the mini-patch sampler in two phases are 165888 and 163840, respectively. Thus, the maximum GPU memory consumption for hybrid NIR models comes from the first phase of optimization.
}
\setlength{\cmidrulekern}{0.25em} 
\begin{tabular}{ccccccc}
\toprule
\multirow{2}{*}{Category}              & \multicolumn{1}{c}{Experiment}  & \multicolumn{2}{c}{(1) Mindboggle101} & \multicolumn{2}{c}{(2) OASIS}   &   \\
\cmidrule(lr){3-4} \cmidrule(lr){5-6}
                                        & Method / Metrics & $\displaystyle DSC_s^{(1mm)}~(\uparrow)$    & $\mathbf{J}_{\leq 0}~(\downarrow)$    & $\displaystyle DSC_v~(\uparrow)$    & $\mathbf{J}_{\leq 0}~(\downarrow)$   & GPU Memory (MB)                        \\
\midrule
\multirow{2}{*}{Learning-based}         & VoxelMorph     & 0.7598           & 2.46e-04              & 0.8204            & 1.96e-4               &$~~8036\mid3837$   \\
                                        & SYM\_Net       & 0.7708           & 7.65e-06              & 0.8265            & 4.60e-06              &$10565\mid3031$    \\
\midrule
\multirow{3}{*}{Optimization-based}     & NiftyReg       & 0.7874           & 1.01e-03              & 0.8234            & 1.28e-03              & -                  \\
                                        & SyN            & 0.7822           & 4.40e-06              & 0.8371            & 6.38e-06              & -                  \\
                                        & Grid          & 0.7110           & 9.38e-04              & 0.7617            & 1.08e-03              & 4981                  \\
\midrule
\multirow{2}{*}{\textbf{Ours}}          & NIR-H        & 0.7809           & 1.31e-04              & 0.8274            & 1.62e-04              & 3341           \\
                                        & NIR-H-Diff   & \textbf{0.7904}  & \textbf{1.11e-06}     & \textbf{0.8382}            & \textbf{4.55e-06}     & 3177           \\
                                        
\bottomrule
\end{tabular}
\label{tab:results_comp}
\end{table*}
\begin{figure*}
\centering
     \begin{subfigure}[b]{\textwidth}
         \centering
         \includegraphics[width=\textwidth]{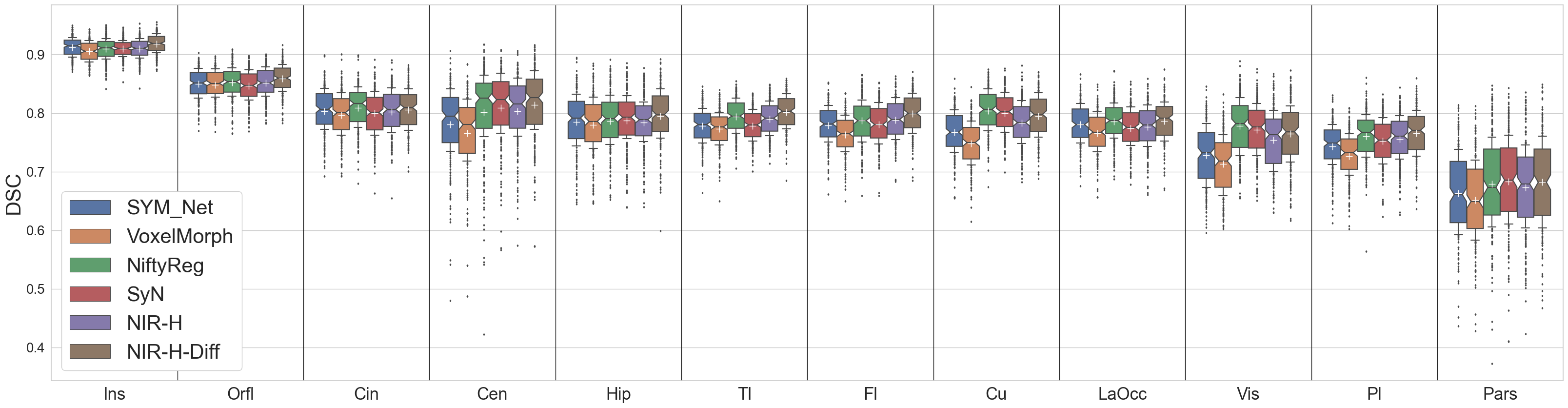}
         \caption{\textbf{Experiment (1)}. We group all 31 structures into 12 groups: Cin (caudal anterior cingulate, rostral anterior cingulate, isthmus cingulate, posterior cingulate),
         Fl (caudal middle frontal, rostral middle frontal, superior frontal),
         Pl (inferior parietal, superior parietal, supramarginal),
         Tl (inferior temporal, middle temporal, superior temporal, transverse temporal),
         Orfl (lateral orbitofrontal, medial orbitofrontal),
         LaOcc (lateral occipital),
         Cen (postcentral, precentral, paracentral),
         Cu (cuneus, precuneus),
         Pars (pars opercularis, pars orbitalis, pars triangularis),
         Hip (entorhinal, parahippocampal),
         Vis (lingual, fusiform, pericalcarine).}
         \label{fig:mind_box}
     \end{subfigure}
     \vfill
     \vspace{1em}
     \begin{subfigure}[b]{\textwidth}
         \centering
         \includegraphics[width=\textwidth]{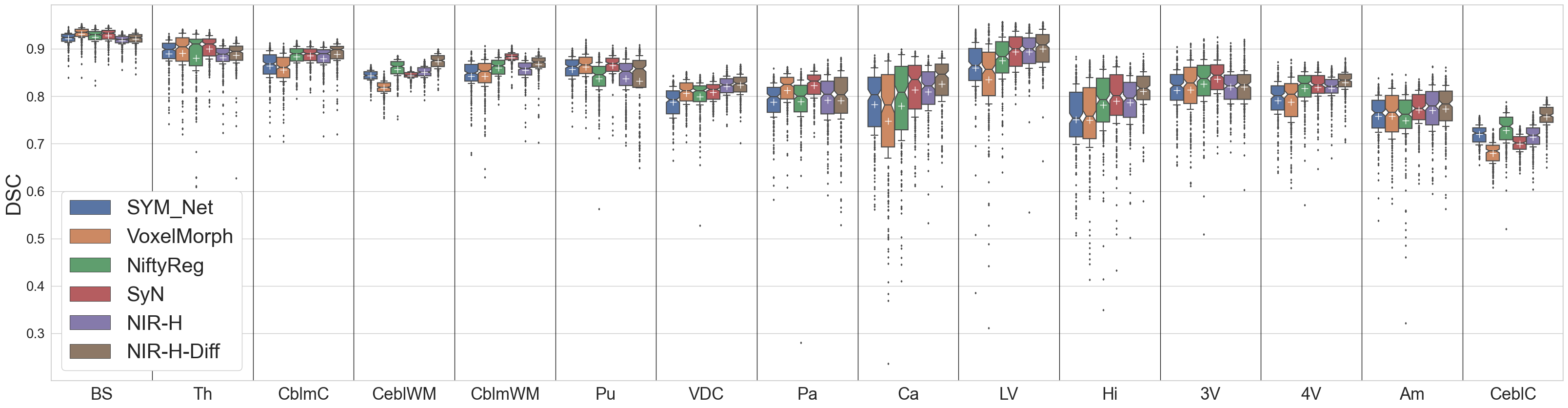}
         \caption{\textbf{Experiment (2).} The abbreviations above indicate: brain stem (BS), thalamus (Th),  cerebellum cortex (CblmC), cerebral white matter (CeblWM), cerebellum white matter (CblmWM), putamen (Pu), Ventral-DC (VDC), Pallidum (Pa), Caudate (Ca), Lateral Ventricle (LV), Hippocampus (Hi), 3rd Ventricle (3V), 4th Ventricle (4V), Amygdala (Am), and
         Cerebral Cortex (CeblC).}
         \label{fig:oasis_box}
     \end{subfigure}
\caption{\textbf{Boxplots of Dice's Coefficients for Various Anatomical Structures}. Left and right hemispheres are combined together, e.g. we average two Dice's Coefficients of the left and right pair of anatomical structures into one. The white '+'s in the above boxes indicate the average Dice's Coefficients.}
\label{fig:boxplot}
\end{figure*}
In this section, we will present the registration performance comparison among the baseline methods and our proposed methods. The scores of our proposed methods reported in Tab.~\ref{tab:results_comp} and Tab.~\ref{tab:generability} are achieved by 900-iteration optimization.
Tab.~\ref{tab:results_comp} shows the performance comparisons among all methods in experiment (1) and (2) over three aspects of criteria --- registration accuracy, registration regularity, and maximum GPU memory consumption during optimization and inference. 

Compared with the learning-based methods (VoxelMorph and SYM\_Net), our GPU memory cost for optimization is less than half of the memory that they consume for training. Moreover, the GPU memory consumption of our methods for optimization is close to that of the two learning-based methods for inference. In terms of registration accuracy, our NIR-H and NIR-H-Diff both perform better than the learning-based methods in both experiment (1) and (2). Especially in experiment (1), our advantages in alignment accuracy applied to almost all annotated structure groups as shown in Fig.~\ref{fig:mind_box}. As for the registration regularity, our NIR-H-Diff can also achieve lower $\mathbf{J}_{\leq 0}$ than another learning-based diffeomorphic registration method SYM\_Net ($\mathbf{J}_{\leq 0}$). It can be observed from Tab.~\ref{tab:results_comp} that our performance edge over the learning-based methods get smaller in experiment (2). A possible explanation for this might be that more data are available for training and all data were recruited from the same institution following the similar scanning protocols in OASIS dataset, therefore the generalization issue of the learning-based methods is not fully exposed in this experiment.

Compared with the optimization-based registration methods (NiftyReg and SyN), our NIR-H-Diff and NIR-H can both provide high-accuracy registration performance but only NIR-H-Diff achieve the the top performance in terms of registration regularity. NIR-H is not comparable with diffeomorphic registration method, i.e., SyN, in the metric of $\mathbf{J}_{\leq 0}$. but it only generates about $1/10$ folds in deformation fields compared with NiftyReg. Fig.~\ref{fig:mind_box} and Fig.~\ref{fig:oasis_box} present a closer inspection of registration accuracy of methods in comparison, from which we can tell that our proposed method can achieve the best performance in 9 out of 12 structure groups in experiment (1) and 9 out of 15 structure groups in experiment (2). Another important criterion to assess the optimization-based methods is the performance relationship with optimization duration, which will be discussed in Sec.~\ref{subsec:results_time_vs_performance}.

Compared with Grid, our NIR-H is significantly better in terms of registration accuracy (>0.6 greater in $\displaystyle DSC$) and regularity ($\approx$10x smaller in $\mathbf{J}_{\leq 0}~(\downarrow)$), consuming less GPU memory. Because Grid and NIR-H share the similar optimization process but mainly differ in the ways to describe the deformation fields, the significant advantage of our proposed method may suggest the effectiveness of neural fields in modeling the deformation fields.

\begin{table}[]
\centering
\caption{
\textbf{Registration Performance Comparison on Experiment (3).} 3 target volumes from Mindboggle101 dataset and 20 moving volumes from OASIS dataset are randomly selected to conduct the cross-dataset image registration experiments. The learning-based methods are trained with the training set of Mindboggle101 dataset.
}
\begin{tabular}{llcc}
\toprule
Category    & Method                  & $\displaystyle SSIM ~(\uparrow)$          & $\mathbf{J}_{\leq 0}~(\downarrow)$                  \\ \midrule
\multirow{2}{*}{\begin{tabular}[c]{@{}l@{}}Learning-\\ based\end{tabular}}  & VoxelMorph    & 0.7238            & 2.94e-04              \\
                                                                            & SYM\_Net      & 0.7543            & 3.06e-05              \\ 
\midrule
\multirow{2}{*}{\textbf{Ours}}                                              & NIR-H       & 0.8408            & 2.30e-04              \\
                                                                            & NIR-H-Diff  & \textbf{0.8530}   & \textbf{2.28e-06}     \\ \bottomrule
\end{tabular}
\label{tab:generability}
\end{table}

What's more, Tab.~\ref{tab:generability} compares the performance of NIR-H, NIR-H-Diff and two learning-based methods in experiment (3). As the moving volumes are healthy scans from the Mindboggle101 dataset and 20 target scans with Alzheimer’s disease come from the OASIS dataset, the results in experiment (3) might suggest the robustness of an algorithm against modest domain shift. It is apparent from Tab.~\ref{tab:generability} that, compared with our proposed method, the learning-based methods learned from the one dataset cannot as well generalize to pair of images coming from different datasets with different health status. To be specific, our NIR-H-Diff method can achieve almost 0.1 higher and more than 10x fewer folds \cite{shen2019networks} in predicted deformation fields, compared with the state-of-the-art learning-based methods SYM\_Net.

\subsection{Qualitative Comparisons with Baselines}
\label{subsec:results_comp_qualitative}
\begin{figure*}
\centering
     \begin{subfigure}[b]{0.85\textwidth}
         \centering
         \includegraphics[width=\textwidth]{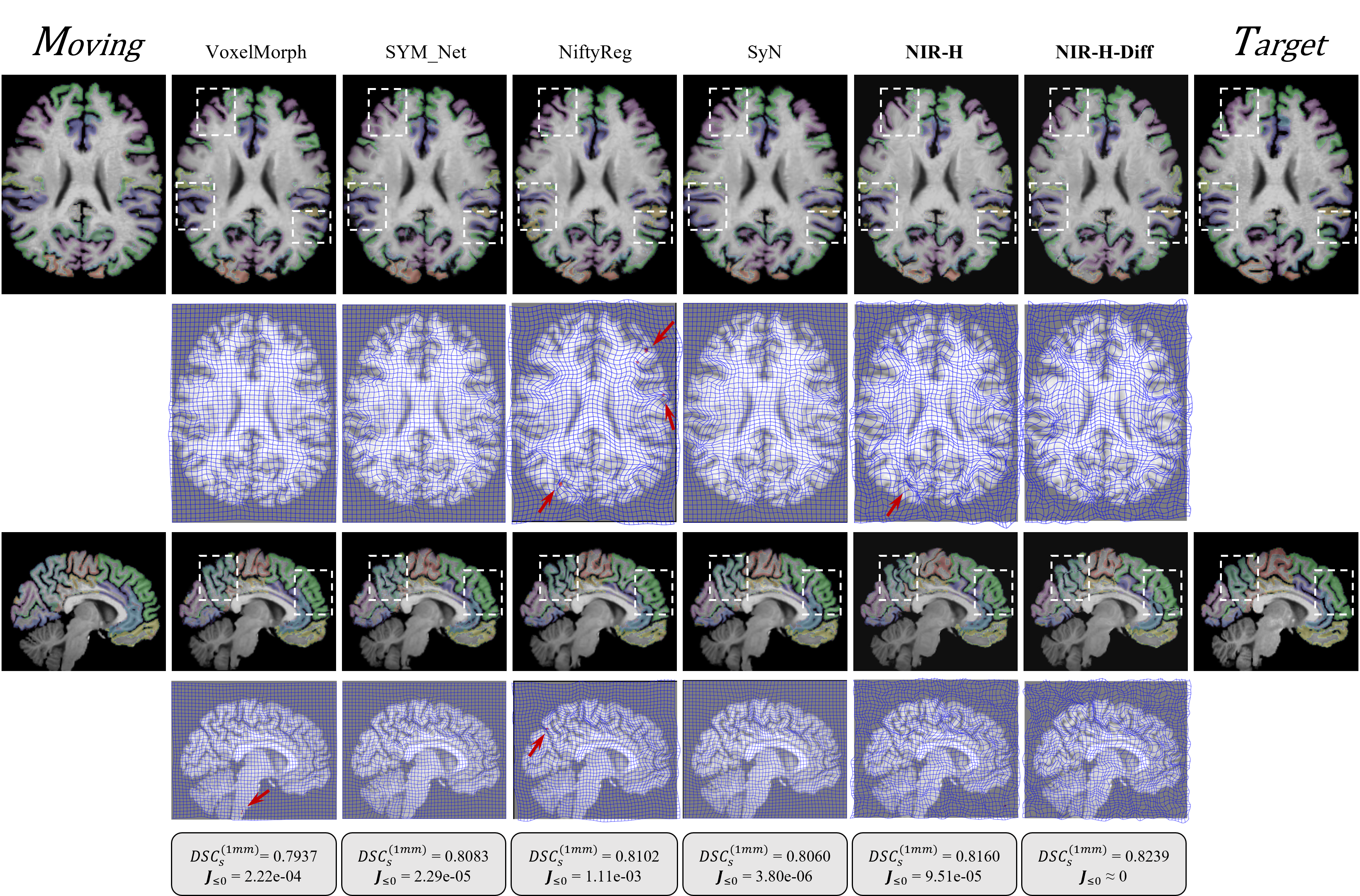}
         \caption{Experiment (1)}
         \label{fig:mind_qualitative_comp}
     \end{subfigure}
     \vfill
     \begin{subfigure}[b]{0.85\textwidth}
         \centering
         \includegraphics[width=\textwidth]{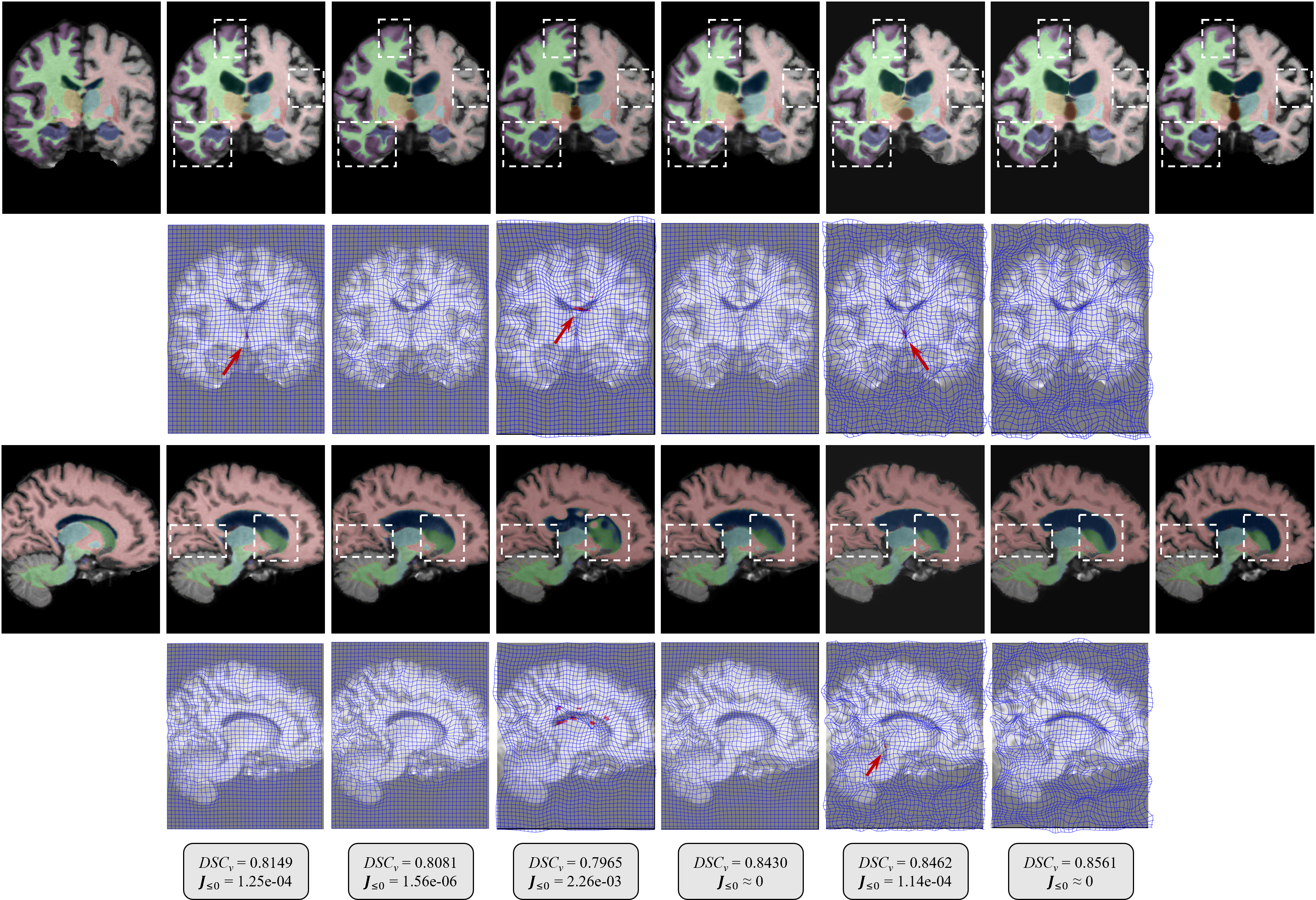}
         \caption{Experiment (2)}
         \label{fig:oasis_qualitative_comp}
     \end{subfigure}
\caption{\textbf{Qualitative Registration Performance Comparison of Different Methods}. The models in qualitative comparison are VoxelMorph, SyM\_Net, NiftyReg, NIR-H and NIR-H-Diff. In the above plots, we present two volume pairs from experiment (1) and experiment (2) in two views. The warped volumes generated by different methods are overlapped with the warped structures which are indicated by colors. The key differences in registration quality of different methods are highlighted by the white dotted boxes. The deformation fields are illustrated by the downsized deformed grid in blue and the regions with negative jacobian determinant are colored in red. The last row in Fig.~(a) and (b) are the quantitative performance of different registration methods on that image pair. If the $\mathbf{J}_{\leq 0}$ is less than $1e-06$, we take it as $\approx 0$.}
\label{fig:qualitative_comp}
\end{figure*}
Fig.~\ref{fig:qualitative_comp} presents the qualitative comparisons between our proposed and baseline methods.
From Fig.~\ref{fig:qualitative_comp}, we can see that all registration methods in comparison can well align the subcortical regions and what makes different lies in the cortex regions within the white dotted boxes. Since the aligned structures in experiment (2) are mostly located in the subcortical regions, the gap between our methods and the baseline methods in experiment (2) should not be as significant as that in experiment (1), which agrees with the results in Tab.~\ref{tab:results_comp} and Fig.~\ref{fig:boxplot}.
Parameterized by a neural field with well-behaved derivatives, the velocity flow is continuously differentiable in NIR-H-Diff. Therefore our registration method doesn't require some explicit smooth operations over the velocity flow to guarantee diffeomorphic transformation. This might explain why NIR-D-Diff can generate deformations with more diverse magnitudes and local orientations on the brain surface, but simultaneously topology preserving is barely touched. 
\subsection{Ablation Study}
\label{subsec:results_ablation_study}

\subsubsection{Influence of Coordinate Samplers}
\begin{table*}[]
\centering
\caption{
\textbf{Registration Performance Differences Resulting from Coordinate Samplers.} The below shows the comparisons of diffeomorphic NIR frameworks optimized via the downsize sampler and mini-patch sampler as well as hybrid diffeomorphic NIR. The comparisons are based on the registration accuracy ($\displaystyle DSC_v$), registration regularity ($\mathbf{J}_{\leq 0}$) and converge speed (Iteration). The iteration number of NIR-H-Diff is that of the second phase of optimization. This table supports the qualitative comparisons of different coordinate samplers as shown in Fig.~\ref{fig:cs_performance_comparison}.
}
\begin{tabular}{lccccccccc}
\toprule
Metrics            & \multicolumn{4}{c}{$\displaystyle DSC_v~(\uparrow)$}       & \multicolumn{4}{c}{$\mathbf{J}_{\leq 0}~(\downarrow)$}    & GPU Memory               \\ 
\cmidrule(lr){2-5} \cmidrule(lr){6-9}
Method / Iteration & 100    & 300    & 600    & 900    & 100      & 300      & 600      & 900       & (MB)      \\ \midrule
NIR-D-Diff       & 0.8250 & \textbf{0.8358} & 0.8369 & 0.8371 & 3.69e-06 & 6.19e-05 & 1.71e-04 & 2.74e-04  & 3177 \\
NIR-P-Diff       & 0.7363 & 0.7936 & 0.8297 & 0.8059 & \textbf{0} & \textbf{1.28e-07} & 3.08e-06 & 5.72e-06  & \textbf{3149} \\ \midrule
NIR-H-Diff       & \textbf{0.8281} & 0.8347 & \textbf{0.8370} & \textbf{0.8382} & 4.57e-06 & 2.00e-06 & \textbf{2.05e-06} & \textbf{4.55e-06}  & 3177 \\ \bottomrule
\end{tabular}
\label{tab:comp_among_cs}
\end{table*}
In Sec.~\ref{subsubsection:coords_sampling}, we have visually compared the effect of different coordinate samplers on the registration performance, here we present the quantitative comparison to support our analysis and designs. 
Tab.~\ref{tab:comp_among_cs} shows the diffeomorphic registration performance differences in experiment (2), resulting from the selections of coordinate samplers. In the table, there is a clear trend that registration accuracy of both NIR-D-Diff and NIR-P-Diff improves over time, but registration regularity deteriorates. Furthermore, the results indicate that NIR-D-Diff can more quickly converge to the higher registration accuracy than NIR-P-Diff. On the other hand, NIR-P-Diff is able to maintain very small $\mathbf{J}_{\leq 0}$ throughout the optimization process, whereas NIR-D-Diff fails.
Based on the above observations, we proposed NIR with a hybrid sampling scheme whose registration performance in Tab.~\ref{tab:comp_among_cs} justifies our design. NIR-H-Diff barely costs extra memory for optimization and outperforms NIR-D-Diff and NIR-P-Diff in registration accuracy and regularity. It should be clarified that in Tab.~\ref{tab:comp_among_cs}, the iteration number for NIR-H-Diff refers to the second phase of optimization, which means, NIR-H-Diff requires 200 more iterations than the other two methods. Nevertheless, our design still meets our expectations in terms of developing a method that can converge quickly to a decent registration results with high accuracy and good topology preserving in an efficient way. 

\subsubsection{Influence of Regularization Weight $\lambda_{reg}$}
\begin{table}[]
\centering
\caption{
\textbf{Influence of Regularity Weight $\lambda_{jdet}$}. The effect of $\lambda_{jdet}$ in the balance of $\displaystyle DSC_v$ and $\mathbf{J}_{\leq 0}$ can be observed on NIR-D-Diff and NIR-P-Diff. However, by merely adjusting the scale of $\lambda_{jdet}$, both methods cannot outperform NIR-H-Diff in terms of registration accuracy and regularity.}
\begin{tabular}{lccc}
\toprule
Method                                            & $\lambda_{jdet}$ & $\displaystyle DSC_v~(\uparrow)$   & $\mathbf{J}_{\leq 0}~(\downarrow)$    \\ \midrule
\multicolumn{1}{c}{\multirow{3}{*}{NIR-D-Diff}} & 100    & 0.8371 & 2.74e-04 \\
\multicolumn{1}{c}{}                              & 1000   & 0.8332 & 1.68e-05 \\
\multicolumn{1}{c}{}                              & 10000  & 0.8193 & 4.99e-06 \\ \midrule
\multirow{2}{*}{NIR-P-Diff}                     & 10     & 0.8330 & 2.63e-05  \\
                                                  & 100    & 0.8332 & 5.72e-06 \\ \midrule
NIR-H-Diff                                      & 100    & \textbf{0.8382} & \textbf{4.55e-06} \\ \bottomrule
\end{tabular}
\label{tab:lambda_ablation_study}
\end{table}
We then investigate whether simply adjusting the weight of regularization term can reach the performance comparable to what NIR with a hybrid coordinate sampler can achieve. All the diffeomorphic registration methods in Tab.~\ref{tab:lambda_ablation_study} are conducted under the setting of experiment (2).
Since we are not satisfied with the performance of NIR-D-Diff in terms of registration regularity, we increase $\lambda_{reg}$ used for optimizing NIR-D-Diff. On the contrary, we decrease $\lambda_{reg}$ used for optimizing NIR-P-Diff in the hope of improving registration accuracy. It turns out, when NIR-D-Diff achieves the comparable $\mathbf{J}_{\leq 0}$ to NIR-H-Diff, the scale of $\lambda_{reg}$ needs to be 100 times larger than the original value, and its $\displaystyle DSC$ is greatly decreased. Surprisingly, decreasing $\lambda_{reg}$ for NIR-P-Diff cannot improve the registration accuracy as expected but indeed harm the registration regularity. In a word, the results in Tab.~\ref{tab:lambda_ablation_study} support that NIR with a hybrid coordinate sampler is a more effective way to balance the registration accuracy and regularity, compared with simply adjusting the regularization weight.
\subsection{Optimization Duration v.s. Registration Performance}
\label{subsec:results_time_vs_performance}

\begin{figure}
\centering
\includegraphics[width=0.49\textwidth]{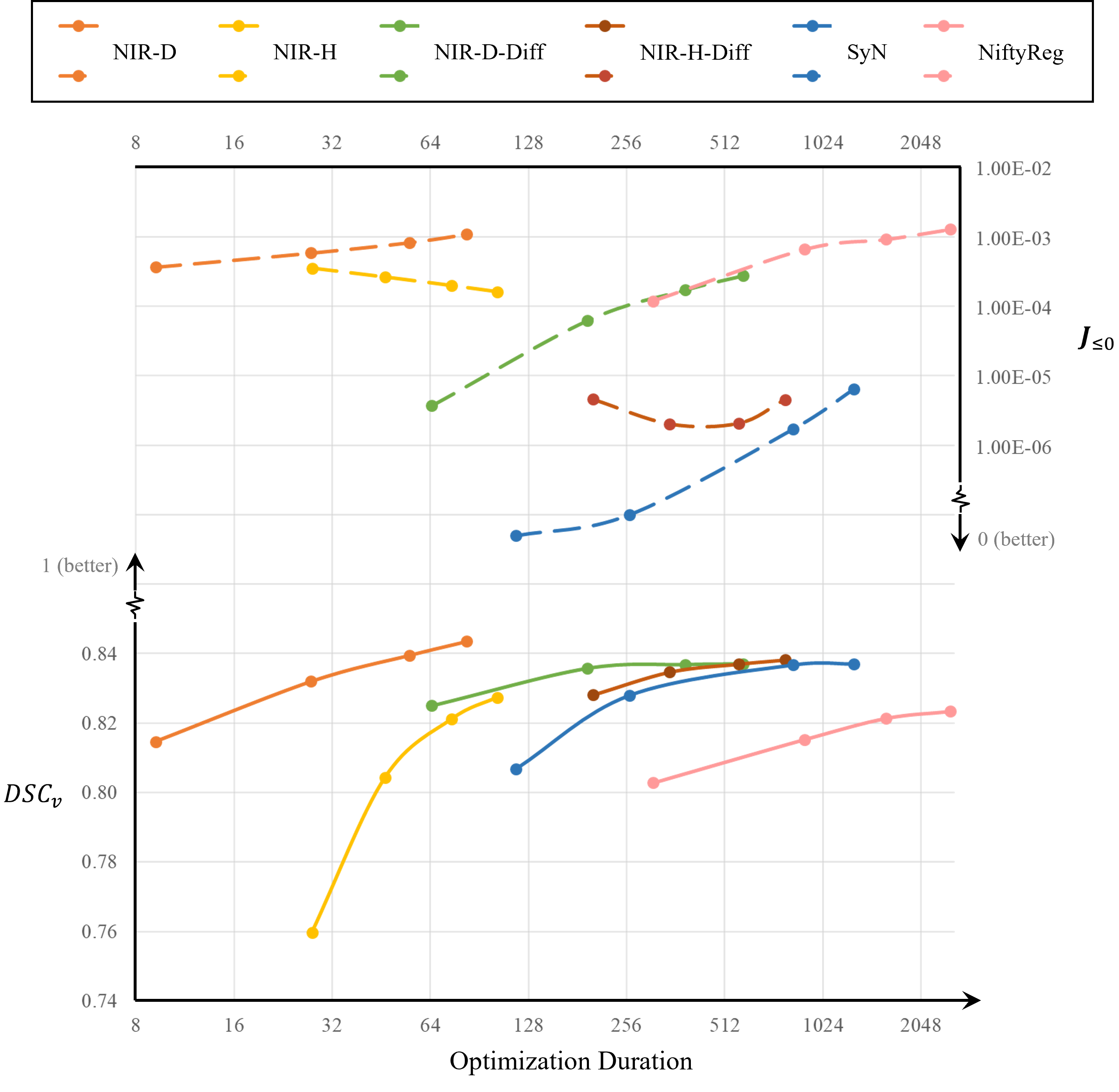}
\caption{\textbf{Optimization Duration v.s. Registration Performance in Experiment (2)}. The solid and dotted curves respectively illustrate the change of registration accuracy and regularity over optimization duration. In the bottom half of this plot, the higher a solid curve goes, the better registration accuracy it indicates. While in the top half this plot, the lower a dotted curve goes, the better registration regularity it reflects. Thus, visually speaking, a method is preferred if its solid and dotted curves get close over time.}
\label{fig:time_vs_performance}
\end{figure}

Fig.~\ref{fig:time_vs_performance} presents the relationship between registration performance in experiment (2) and optimization duration of six optimization-based registration methods, four of which are our proposed methods and the other two are SyN and NiftyReg. 

As for our proposed methods, we evaluate their registration performance at \{100, 300, 600, 900\} optimization iterations. To finish 100-iteration optimization, the displacement field based NIR methods take about 9s and the diffeomorphic NIR methods take about 64s. It needs to be clarified that the optimization iteration of NIR-H and NIR-H-Diff counts from the start of the second-phase optimization. As for SyN, we evaluate the registration performance when setting the maximum optimization iteration as \{8, 4, 2\}, \{20, 10, 5\}, \{60, 30, 15\} and \{100, 50, 25\}
at each level, and the average of corresponding optimization time is about 117s, 261s, 829s and 1273s. In terms of NiftyReg, we evaluate the registration performance with the maximum optimization iteration as \{120, 60, 30\}, \{400, 200, 100\}, \{800, 400, 200\} and \{1200, 600, 300\}
at each level, and the average optimization time is approximately 309s, 901s, 1638s and 2521s.

Among all methods in comparison, NIR-D has the fastest converge speed and the highest $\displaystyle DSC_v$ (0.8435), but it also achieves the highest $\mathbf{J}_{\leq 0}$ (1.08e-03). NIR-H can mitigate the registration regularity issue of NIR-D at the cost of lower registration accuracy. From Fig.~\ref{fig:time_vs_performance}, we can see that NIR-H has the potential to get improved in both $\displaystyle DSC_v$ and $\mathbf{J}_{\leq 0}$ if the optimization duration is extended. 
NiftyReg performs relatively bad at converge speed, registration accuracy as well as registration regularity. Because GPU acceleration has been disabled by the official implementation of NiftyReg, optimizing with $LNCC$ similarity becomes so time-consuming that it is even slower than those methods supporting diffeomorphic transformations. 

NIR-D-Diff is capable of reaching a very decent registration accuracy ($\displaystyle DSC_v \geq 0.83$) in a short amount of time ($\approx200s$), but the registration regularity degrade as the number of optimization iterations grows. NIR-H-Diff is proposed to achieve a better balance between the registration accuracy and regularity. While achieving the similar $\displaystyle DSC_v$ as NIR-D-Diff, NIR-H-Diff can obtain considerably greater regularity of deformation fields, i.e., $\mathbf{J}_{\leq 0}$ stays lower than 5e-06 during optimization.
SyN shows very strong performance in experiments (2), especially in terms of registration regularity. Despite this, it is demonstrated in Fig.~\ref{fig:time_vs_performance} that our approaches have two main advantages over SyN. First, optimized for the similar duration, our NIR-D-Diff and NIR-H-Diff can achieve higher $\displaystyle DSC_v$ scores than SyN. Second, SyN gets significantly worse registration regularity as optimization iterations in the finer scale get increased, and ends with a higher $\mathbf{J}_{\leq 0}$ than our NIR-H-Diff.

\section{Limitations and Future Directions}
\label{sec:future_works}

One major limitation of NIR is its running time. Although significantly faster than traditional optimization-based methods, it is still much slower than learning-based methods. There are a few potential approaches to address this limitation.
First, we can design an adaptive coordinate sampler that samples coordinates sparsely in easy-to-align regions, but densely in the regions with large alignment errors. 
Second, NIR can be used in conjunction with a learning-based method in a two-step approach, using the learning-based method to generate an initial registration, followed by fine-tuning through NIR. 
Third, neural fields can also be integrated into a learning-based framework \cite{sun2022topology,park2019deepsdf,zheng2021deep}, where the coordinate-based MLPs and an embedding layer are learned from the training data. During inference, the parameters of coordinate-based MLPs are fixed and merely a latent code associated with the test data is optimized. 

In addition, how to introduce surface registration into our image registration framework is a topic worth exploring. NIR establishes correspondence between image pairs to match voxel intensities.  It is agnostic to anatomic structures within the images and thus does not always lead to semantically meaningful registrations.
One future direction in this regard is to optimize both intensity and shape similarities between two images. Since shape registration can also be realized via neural fields as we showed previously \cite{sun2022topology}, neural fields provide a promising approach to unify both intensity-based and shape-based registrations within the same framework.  

\section{Conclusions}
\label{sec:conclusions}

We presented a new optimization-based framework, named NIR, for deformable image registration. NIR uses coordinate-based MLPs with Fourier position encoding and sinusoidal action functions to model deformation vector fields, and leverages the full power of existing deep learning toolboxes to solve the optimization efficiently.   
%
%

We presented several options of running NIR, depending on the type of registrations (deformable or diffeomorphic) and the speed requirement:  
a) \textbf{NIR-D}: the fastest displacement-based deformable registration method;
b) \textbf{NIR-H}: a rapid displacement-based deformable registration method with a better registration regularity compared to NIR-D;
c) \textbf{NIR-D-Diff}: a diffeomorphic registration method with a good registration regularity;
and 
d) \textbf{NIR-H-Diff}: a slightly slower diffeomorphic registration method with the best registration regularity. 

We compared our methods with several benchmarks on two brain MRI datasets and show that our  methods  achieve state-of-the-art performances in both registration accuracy and regularity.
Compared to the traditional optimization-based methods, our methods achieve competitive results with significantly shorter running time.
Compared to the learning-based methods, our methods show significantly better generalization ability. 
%
%
%
%

Modeling deformation with neural fields offers some major advantages - can model complex deformations with the expressive power of deep neural nets, and can solve optimization efficiently with existing deep learning toolboxes. We believe it offers an appealing alternative for solving the long-standing image registration problem.




{
    \small
    \bibliographystyle{ieee_fullname}
    \bibliography{macros,main}
}



\end{document}